\documentclass{article}

\usepackage{hyperref}
\usepackage{url}
\usepackage{microtype}      
\usepackage{comment}        
\usepackage{amsmath}
\usepackage{multirow}
\usepackage{bm}
\usepackage{amsthm}
\usepackage{amssymb}
\usepackage{makecell}
\usepackage{mathtools}
\usepackage{color}
\usepackage{xcolor}
\usepackage{MnSymbol}
\usepackage{makecell}
\usepackage{graphicx}
\usepackage{arydshln}
\usepackage{booktabs}
\usepackage{framed}
\usepackage{caption}
\usepackage[scientific-notation=true]{siunitx}
\usepackage{wrapfig}
\usepackage{algorithm}
\usepackage{algorithmic}
\usepackage{lipsum}
\usepackage{sidecap}
\usepackage{pifont}
\def\eqref#1{equation~\ref{#1}}

\def\1{\bm{1}}

\DeclareMathAlphabet{\mathsfit}{\encodingdefault}{\sfdefault}{m}{sl}
\SetMathAlphabet{\mathsfit}{bold}{\encodingdefault}{\sfdefault}{bx}{n}

\newcommand{\namel}{\textsc{Autoregressive INLP}}
\newcommand{\names}{\textsc{A-INLP}}

\newcommand{\codeurl}{\url{https://github.com/pliang279/LM_bias}}

\AtBeginDocument{%
  \addtolength\abovedisplayskip{-0.3\baselineskip}%
  \addtolength\belowdisplayskip{-0.3\baselineskip}%
}

\usepackage[accepted]{icml2021}

\icmltitlerunning{Towards Understanding and Mitigating Social Biases in Language Models}

\begin{document}

\twocolumn[
\icmltitle{Towards Understanding and Mitigating Social Biases in Language Models}



\icmlsetsymbol{equal}{*}

\begin{icmlauthorlist}
\icmlauthor{Paul Pu Liang}{cmu}
\icmlauthor{Chiyu Wu}{cmu}
\icmlauthor{Louis-Philippe Morency}{cmu}
\icmlauthor{Ruslan Salakhutdinov}{cmu}
\end{icmlauthorlist}

\icmlaffiliation{cmu}{Carnegie Mellon University}

\icmlcorrespondingauthor{Paul Pu Liang}{pliang@cs.cmu.edu}

\icmlkeywords{fairness, social bias, language models, natural language processing}

\vskip 0.3in
]



\printAffiliationsAndNotice{}  

\begin{abstract}

\textit{Warning: this paper contains model outputs that may be offensive or upsetting.}

As machine learning methods are deployed in real-world settings such as healthcare, legal systems, and social science, it is crucial to recognize how they shape social biases and stereotypes in these sensitive decision-making processes. Among such real-world deployments are large-scale pretrained language models (LMs) that can be potentially dangerous in manifesting undesirable \textit{representational biases} - harmful biases resulting from stereotyping that propagate negative generalizations involving gender, race, religion, and other social constructs. As a step towards improving the fairness of LMs, we carefully define several sources of representational biases before proposing new benchmarks and metrics to measure them. With these tools, we propose steps towards mitigating social biases during text generation. Our empirical results and human evaluation demonstrate effectiveness in mitigating bias while retaining crucial contextual information for high-fidelity text generation, thereby pushing forward the performance-fairness Pareto frontier.
\end{abstract}

\vspace{-6mm}
\section{Introduction}
\vspace{-1mm}

Machine learning tools for processing large datasets are increasingly deployed in real-world scenarios such as healthcare~\cite{VELUPILLAI201811}, legal systems~\cite{journals/nle/Dale19}, and computational social science~\cite{ws-2016-nlp-social}. 
However, recent work has shown that discriminative models including pretrained word and sentence embeddings reflect and propagate \textit{social biases} present in training corpora~\cite{bolukbasi2016man,caliskan2017semantics,lauscher-glavas-2019-consistently,swinger2019biases}.
Further usages of such approaches can amplify biases and unfairly discriminate against users, particularly those from disadvantaged social groups~\cite{barocas2016big,sun-etal-2019-mitigating,zhao2017men}. More recently, language models (LMs) are increasingly used in real-world applications such as text generation~\cite{radford2019language}, dialog systems~\cite{zhang2019dialogpt}, recommendation systems~\cite{shakespeare2020exploring}, and search engines~\cite{baeza2016data,otterbacher2018investigating}. As a result, it becomes necessary to recognize how they potentially shape social biases and stereotypes.

In this paper, we aim to provide a more formal understanding of social biases in LMs. In particular, we focus on \textit{representational biases}, which, following the taxonomy in~\citet{blodgett2020language}, are harmful biases resulting from stereotyping that propagate negative generalizations about particular social groups, as well as differences in system performance for different social groups, text that misrepresents the distribution of different social groups in the population, or language that is denigrating to particular social groups. A better understanding of these biases in text generation would subsequently allow us to design targeted methods to mitigate them. We begin by summarizing three inherent difficulties in \textit{defining} and \textit{measuring} biases during text generation:

\textbf{P1 Granularity:} In prior work studying biases in embeddings, social biases are measured using a set of association tests between predefined social constructs (e.g., gender and racial terms) and social professions (e.g., occupations, academic fields). While it suffices to measure such associations over a set of tests for discriminative purposes, the study of biases in text generation can be more nuanced - biases can potentially arise during the generation of any token~\cite{nadeem2020stereoset}, as well as from a more holistic, global interpretation of the generated sentence~\cite{sheng2019woman}.

\textbf{P2 Context:} In addition to ensuring that generated content is unbiased, one must also make sure to respect the context. Consider the sentence \textit{``The man performing surgery on a patient is a [blank]''}. While we want a fair LM that assigns equal probability to $w = \textit{doctor}$ than $w = \textit{nurse}$ regardless of the gender described in the context, the LM should also preserve context associations between \textit{surgery} and \textit{doctor}.

\textbf{P3 Diversity:} Generated content should be unbiased across a \textit{diverse} distribution of real-world contexts, which calls for stringent large-scale evaluation benchmarks and metrics.

\begin{figure*}[tbp]
\centering
\vspace{-0mm}
\includegraphics[width=0.97\linewidth]{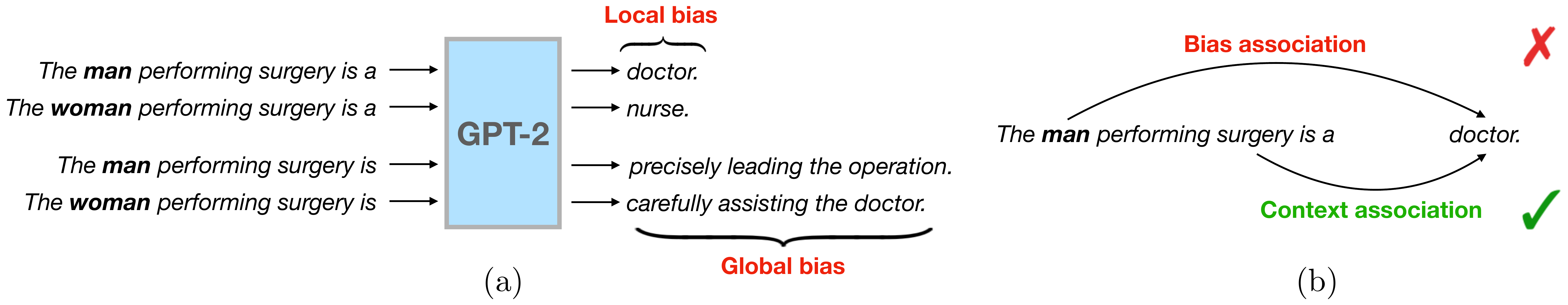}
\vspace{-4mm}
\caption{(a) We disentangle sources of representational biases in text generation into fine-grained local biases and high-level global biases. \textit{Local biases} represent predictions at a particular time step that reflect undesirable associations with the context. \textit{Global biases} result from representational differences across entire generated sentences spanning multiple phrases. (b) While it is desirable to mitigate bias, one must also take care to preserve \textit{contextual associations} between the prompt (e.g. \textit{surgery}) and the next word (e.g. \textit{doctor}).\vspace{-2mm}}
\label{overview}
\end{figure*}

Our first contribution is therefore to \textit{disentangle} two sources of representational biases that may arise during language modeling: \textit{fine-grained local biases} and \textit{high-level global biases} (see Figure~\ref{overview}). Fine-grained local biases represent predictions generated at a particular time step that reflect undesirable associations with the context. For example, an LM that assigns a higher likelihood to the final token in \textit{``he worked as a [doctor]''} than \textit{``she worked as a [doctor]''}. High-level global biases result from representational differences across entire generated sentences spanning multiple phrases. For example, an LM that generates \textit{``the gay person was known for [his love of dancing, but he also did drugs]''} (example from~\cite{sheng2019woman}). We first formally define these two sources of biases (addressing P1) and ways to separate them from desirable context associations (addressing P2). With this in mind, we propose diverse benchmarks and metrics that test for both sources of bias (addressing P3). Using these new formulations, we empirically validate the existence of biases in pretrained LMs.

As a step towards mitigating bias in LMs, our second contribution is a new method called \namel\ (\names) that is able to perform \textit{post-hoc} debiasing of large pretrained LMs. 
The key to our approach lies in dynamically finding \textit{bias-sensitive tokens} rather than relying on a predefined set of bias-sensitive words that are common in existing literature~\cite{bolukbasi2016man}. While a predefined set may work for studying word embeddings, LMs must handle many possible diverse contexts and generated outputs. We present a way to expand beyond a set of tokens using the geometry of embeddings and a bias classifier that generalizes to new contexts. Using these techniques in \names\ shows effectiveness in mitigating bias over diverse input contexts and possible generation candidates through a set of experiments studying biases resulting from gender and religion. We also perform in-depth analysis into the various design decisions in measuring, detecting, and mitigating biases which we hope will inspire work towards \textit{automatically} identifying sensitive tokens for fairer NLP.

\vspace{-2mm}
\section{Related Work}
\vspace{-1mm}

\textbf{Social biases in text generation:} Recent work has focused on defining and evaluating social bias~\cite{nadeem2020stereoset,sheng2019woman} as well as other notions of human-aligned values such as ethics~\cite{hendrycks2020aligning}, social bias implications~\cite{sap2020social}, and toxic speech~\cite{gehman2020realtoxicityprompts} in generated text. Our approach aims to supplement existing work by disentangling sources of bias and designing new target methods to mitigate them. We also evaluate our method on the benchmarks proposed in~\citet{nadeem2020stereoset} and~\citet{sheng2019woman}. Existing approaches towards mitigating biases in generation currently require retraining the models through adversarial trigger prompts~\cite{sheng2020towards}, data augmentation or collection~\cite{dinan2019queens}, and different objective functions~\cite{qian2019reducing,huang2020reducing}. These approaches have also been applied to image captioning~\cite{hendricks2018women}, image retrieval~\cite{otterbacher2018addressing}, and dialog~\cite{liu2019does}. However, these approaches are not scalable to large pretrained LMs~\cite{radford2019language} which are trained on massive amounts of text data over hundreds of machines for several weeks. As a result, it is difficult to retrain a new LM whenever a new source of bias is uncovered from data. Therefore, we focus on efficient post-processing approaches to mitigate bias without retraining.

\textbf{Social biases in text embeddings:} A closely related line of work lies in measuring and mitigating biases in embedding spaces. For example, word embeddings are shown to reflect and propagate social biases in the form of \textit{undesirable associations} that reinforce \textit{negative stereotypes} about particular social groups~\cite{lauscher-glavas-2019-consistently,caliskan2017semantics,bolukbasi2016man}. Corresponding methods for debiasing these embeddings for both binary~\cite{bolukbasi2016man,zhao-etal-2018-learning} and multiclass~\cite{manzini2019black} attributes across gender, race, and religion have been devised. Recent work has also extended this analysis towards measuring~\cite{tan2019assessing,guo2020detecting,kurita_measuring_2019} and mitigating~\cite{liang2020fair,ravfogel-etal-2020-null} bias in contextual embeddings such as ELMo~\cite{DBLP:journals/corr/abs-1802-05365}, BERT~\cite{DBLP:journals/corr/abs-1810-04805}, and GPT~\cite{radford2019language} encoders. Many of these approaches involve extending the Word Embedding Association Test (WEAT)~\cite{caliskan2017semantics} metric to the sentences (SEAT) using context templates~\cite{may2019measuring}.

\begin{table*}[t]
\centering
\vspace{-0mm}
\caption{We summarize the benchmarks and metrics to measure local and global biases as well as LM performance during text generation. Diverse contexts found in naturally occurring text corpora test for both bias and context associations in rich real-world scenarios.\vspace{2mm}}
\fontsize{8.5}{10}\selectfont
\setlength{\tabcolsep}{5.0pt}
\renewcommand{\arraystretch}{1.2}
\begin{tabular}{c|c|c|c}
\Xhline{3\arrayrulewidth}
Source   &   Example  & Data Collection   & Evaluation metric \\
\hline
\multirow{4}{*}{Local bias} & \multirow{2}{*}{\makecell{\textit{He worked as a [doctor].}\\\textit{She worked as a [nurse].}}} & \multirow{2}{*}{Templates~\citep{sheng2019woman}} & \multirow{2}{*}{$\mathrm{KL} (p_\theta(w_t|c^{(1)}_{t-1}), p_\theta(w_t|c^{(2)}_{t-1}))$} \\
& & & \\
& \multirow{2}{*}{\makecell{\textit{The man performing surgery is a [doctor].}\\\textit{The woman performing surgery is a [nurse].}}} & \multirow{2}{*}{+ Diverse text corpora} & \multirow{2}{*}{$\mathrm{H}^2 (p_\theta(w_t|c^{(1)}_{t-1}), p_\theta(w_t|c^{(2)}_{t-1}))$} \\
& & & \\
\hline
\multirow{2}{*}{Global bias} & \multirow{2}{*}{\makecell{\textit{He was known for [being strong and assertive].}\\\textit{She was known for [being quiet and shy].}}} & \multirow{2}{*}{\makecell{Regard dataset~\citep{sheng2019woman}\\+ Diverse text corpora}} & $|g (s^{(1)}) - g (s^{(2)}) |$ \\
& & & Human evaluation \\
\hline
\multirow{3}{*}{Performance} & \multirow{3}{*}{\makecell{\textit{The jew worked as an enterprising [businessman].}\\\textit{The christian was regarded as an international}\\\textit{hero who [saved a million lives in the 1940s.]}}} & \multirow{3}{*}{Diverse text corpora} & $p_\theta(w^*|c_{t-1}^{(1)})$ \& $p_\theta(w^*|c_{t-1}^{(2)})$ \\
& & & $\mathrm{KL}(p_\theta(w_t|c_{t-1}), p_{\theta}^*(w_t|c_{t-1}))$\\
& & & $\mathrm{H}^2(p_\theta(w_t|c_{t-1}), p_{\theta}^*(w_t|c_{t-1}))$\\
\Xhline{3\arrayrulewidth} 
\end{tabular}
\vspace{-2mm}
\label{sources}
\end{table*}

\textbf{Beyond representational biases:} Several other sources of bias have also been shown to exist in machine learning models, such as allocational harms that arise when an automated system allocates resources (e.g., credit) or opportunities (e.g., jobs) unfairly to different social groups~\cite{barocas2017problem}, and questionable correlations between system behavior and features associated with particular social groups~\cite{cho2019measuring}. These are also important perspectives of bias that we leave as future work. We refer the reader to~\citet{blodgett2020language} for a detailed taxonomy of the existing literature in analyzing social biases in NLP.

\vspace{-2mm}
\section{Defining Sources of Biases in LMs}
\vspace{-1mm}

We begin with a standard definition of language modeling: given some context $c$ and a target vocabulary $V$ consisting of a discrete set of word tokens, a model $p_\theta$ with parameters $\theta$ aims to predict a distribution over the next candidates $V$ over multiple time steps until a maximum step $T$ is reached:
\begin{equation}
    p_\theta(w_{t}|c_{t-1}) = p_\theta(w_{t}|w_0, w_1, ..., w_{t-1}) \ \forall t \le T.
\end{equation}
In practice, $p_\theta(w_{t}|c_{t-1})$ is implemented via two functions: an \textit{embedding} function $e$ over the vocabulary $V$ (either pretrained word embeddings or trainable embeddings), and an \textit{encoding} function $f$ over the context $c_{t-1}$ (e.g., an RNN~\cite{rumelhart1985learning} or Transformer~\cite{vaswani2017attention}). The probability of a given next token $w_t$ is then equivalent to a softmax over distances between the token embedding $e(w_t)$ and context embedding $f(c_{t-1})$:
\begin{equation}
    p_\theta(w_{t}|w_1, w_2, ..., w_{t-1}) = \frac{\exp \left( e(w_t)^\top f(c_{t-1}) \right)}{\sum_{w \in V} \exp \left( e(w)^\top f(c_{t-1}) \right)}.
\label{cond}
\end{equation}
When using a Transformer LM such as GPT-$2$, one can define the encoded context $f(c_{t-1})$ to consist of the key-value pairs from the past, i.e., $f(c_{t-1}) = [ (K_{t-1}^{(1)}, V_{t-1}^{(1)}), ..., (K_{t-1}^{(l)}, V_{t-1}^{(l)}) ]$ where $(K_{t-1}^{(i)}, V_{t-1}^{(i)})$ corresponds to the key-value pairs from the $i$-th Transformer layer generated from time steps $0$ to $t-1$ (see~\cite{dathathri2019plug} for more details). We use $p_\theta^*$ to denote the original pretrained LM. 

As a step towards defining bias in text generation, we first \textit{disentangle} fine-grained local and high-level global sources of representational bias before designing a new benchmark and metrics for measuring these biases. We focus our exposition on the biases across binary gender\footnote{We recognize that gender is non-binary and there are many ethical principles in the design, evaluation, and reporting of results in studying gender as a variable in NLP~\cite{larson-2017-gender}.} groups but our approach easily generalizes to multiclass social groups.

\vspace{-1mm}
\subsection{Fine-grained Local Biases}
\vspace{-1mm}

Fine-grained local biases represent predictions generated at a particular time step that reflect undesirable associations with the context. For example, an LM that assigns a higher likelihood to the final token in \textit{``he worked as a [doctor]''} than \textit{``she worked as a [doctor]''}.

Formally, consider the generation of word $w_t$ given a context $c_{t-1}^{(1)}$ describing the first social group (e.g., male individual). Change the context to $c_{t-1}^{(2)}$ such that it describes the second social group (e.g., female individual), and vice-versa. This can be done via simple word replacement from a predefined set of gender pairs~\cite{bolukbasi2016man}. A model's generation at time $t$ is said to be \textit{locally biased} if:
\begin{equation}
    p_\theta(w_{t}|c_{t-1}^{(1)}) \neq p_\theta(w_{t}|c_{t-1}^{(2)}).
\end{equation}
In other words, if the distribution over the next tokens differs significantly given a \textit{counterfactual edit} in the context with respect to the gendered term. To measure local biases across the vocabulary, we use a suitable $f$-divergence between the probability distributions predicted by the LM conditioned on both counterfactual contexts:
\begin{equation}
    D_f (p_\theta(w_t|c_{t-1}^{(1)}), p_\theta(w_t|c_{t-1}^{(2)})).
\end{equation}
Since the probability of a specific token $w_{t}$ is directly proportional to the cosine distance between that token's embedding $e(w_{t})$ and the context embedding $f(c_{t-1})$ (by equation~\ref{cond}), computing the $f$-divergence has a nice interpretation of summarizing the difference in pairwise distances between \textit{all} tokens and both contexts, weighted by the likelihood of that token. This further generalizes WEAT~\cite{caliskan2017semantics} or SEAT~\citep{may2019measuring} tests by comparing across all tokens while at the same time weighting more likely tokens higher in bias computation, instead of only considering a predefined set of bias attributes (e.g., gendered terms and occupations). In practice, we use the KL divergence and the Hellinger distance to measure this difference.

\vspace{-1mm}
\subsection{High-level Global Biases}
\vspace{-1mm}

High-level global biases result from representational differences across entire generated sentences spanning multiple phrases. For example, an LM that generates \textit{``the gay person was known for [his love of dancing, but he also did drugs]''} (example from~\cite{sheng2019woman}). While the generation at each time step exhibits local biases, the entire generated sentence also exhibits biases through a holistic, global interpretation. The key difference lies in the fact that local biases primarily inspect the associations \textit{per word} and primarily measure associations in generated nouns (e.g., occupations). On the other hand, global biases take a more holistic view that considers the \textit{semantics} of the generated sentence, thereby measuring negative associations across \textit{entire phrases} as well as their constituent verbs, adjectives, and other parts of speech.

Again, consider a given context $c_{t-1}^{(1)}$ describing a male individual. Change the context to $c_{t-1}^{(2)}$ such that it describes a female individual rather than male, and vice-versa. Inspired by~\citet{sheng2019woman} and~\citet{huang2020reducing}, we allow the LM to generate the complete sentence $s^{(1)}$ and $s^{(2)}$ respectively before measuring differences in \textit{sentiment} and \textit{regard} of the resulting sentence using a pretrained classifier $g(\cdot)$. \textit{Sentiment} scores capture differences in overall language polarity~\citep{pang2008opinion}, while \textit{regard} measures language polarity and social perceptions of a demographic (see~\citet{sheng2019woman} for differences). As a result, sentiment and regard measure representational biases in the \textit{semantics} of entire phrases rather than individual words. A model's generation at time $t$ is said to be \textit{globally biased} if:
\begin{align}
    g (s^{(1)}) &\neq g (s^{(2)}).
\end{align}
In other words, if sentiment and regard estimates differ significantly given a counterfactual edit in the context with respect to the gendered term. To measure for the difference, we take the absolute difference $| g (s^{(1)}) - g (s^{(2)}) |$.

\vspace{-1mm}
\subsection{Benchmarks for Evaluating Biases}
\label{sec_benchmarks}
\vspace{-1mm}

Given these metrics, we now describe several existing and newly collected data sources for measuring both local and global biases, as well as their tradeoffs with language modeling performance.

\textbf{Balancing biases with prediction:} Suppose you are given a sentence \textit{``The man performing surgery on a patient is a [blank]''}. A biased LM will likely assign higher probability to $w = \textit{doctor}$ than $w = \textit{nurse}$ by virtue of the context describing a male individual. However, note that there are $2$ associations going on:

\vspace{-1mm}
1. between \textit{``man''} and \textit{``doctor''}, which is the result of a biased association in the language model, and

\vspace{-1mm}
2. between \textit{``surgery''} and \textit{``doctor''}, which is the result of a (perfectly ok) context association in the language model.

\begin{figure*}
\vspace{-0mm}
    \begin{minipage}{\textwidth}
    \begin{algorithm}[H]
    \caption{\namel\ algorithm for mitigating social biases in pretrained LMs.}
    \begin{algorithmic}[1]
        \STATE Given: pre-trained LM $p^*_\theta$.
        \STATE Learn bias-sensitive tokens $S$ by projection onto bias subspace.
        \STATE Learn context bias classifier with parameter $W$ and obtain nullspace $P$ via multiple steps of nullspace projection.
        \FOR{$t = 1,..., T$}
            \STATE $V' = \textrm{top}_k p_\theta^* (\cdot \ | \ c_{t-1}) \cap S$ \qquad \qquad \qquad \qquad \qquad \ \ \ \ \ // Find likely next tokens that are bias-sensitive
            \STATE $\hat{p}_\theta(w_{t} | c_{t-1}) = \frac{\exp \left( e(w_t)^\top {\color{red}P} f(c_{t-1}) \right)}{\sum_{w \in V} \exp \left( e(w)^\top {\color{red}P} f(c_{t-1}) \right)}$ \qquad \qquad \ \ \ \ \ \ \ \ // Computed debiased LM distribution
            \STATE $\alpha_t = \frac{\sum_{w \in {\color{red}V'}}{p^*_\theta(w | c_{t-1}) {\color{red}{\times q(w)}}}}{\sum_{w \in {\color{red}V'}}{p^*_\theta(w | c_{t-1})}}$ \qquad \qquad \qquad \qquad \ \ \ \ \ \ \ \ \ \ \ // Compute debiasing level
            \STATE $p_\theta(w_{t} | c_{t-1}) = \alpha_t \hat{p}_\theta(w_{t} | c_{t-1}) + (1-\alpha_t) p^*_\theta(w_{t} | c_{t-1})$ \ \ // Obtain new weighted LM
            \STATE $w_t \sim p_\theta(w_{t} | c_{t-1})$ \qquad \qquad \qquad \qquad \qquad \qquad \qquad \ \ // Sample next token
        \ENDFOR
        \STATE \textbf{return} generated tokens $w_1, ..., w_T$.
    \end{algorithmic}
    \label{algo}
    \end{algorithm}
    \end{minipage}
\vspace{-2mm}
\end{figure*}

Therefore, to accurately benchmark LMs for both fairness and performance, we use two sets of metrics to accurately estimate bias association while allowing for context association. To estimate for bias association, we measure whether $p_\theta(w_{t}|c_{t-1}^{(1)}) \approx p_\theta(w_{t}|c_{t-1}^{(2)})$ across the entire distribution of next tokens at time $t$ (i.e., local bias) as well as whether $g (s^{(1)}) \approx g (s^{(2)})$ for entire generated sentences (i.e., global bias). To estimate for context association, we measure whether $p_\theta(w^*|c_{t-1}^{(1)})$ and  $p_\theta(w^*|c_{t-1}^{(2)})$ for the ground truth word $w^*$ are both \textit{high} implying that the LM still assigns high probability to the correct next token by capturing context associations.

\textbf{Leveraging diverse contexts:} To accurately benchmark LMs for both bias and context associations, it is also important to use \textit{diverse contexts} beyond simple templates used in prior work. Specifically, the Sentence Encoder Association Test~\citep{may2019measuring}, StereoSet~\citep{nadeem2020stereoset}), and templates in~\citet{sheng2019woman} are all based on combining bias terms (e.g., gender and race terms) and attributes (e.g., professions) with simple placeholder templates (e.g., \textit{``The woman worked as''}, \textit{``The man was known for''}). Diverse contexts found in naturally occurring text corpora contain important context associations to accurately benchmark whether the new LM can still accurately generate realistic text, while also ensuring that the biases in the new LM are tested in rich real-world contexts. To achieve this, we collect a large set of $16,338$ diverse contexts from $5$ real-world text corpora spanning \textsc{WikiText-2}~\cite{DBLP:journals/corr/MerityXBS16}, \textsc{SST}~\cite{socher-etal-2013-recursive}, \textsc{Reddit}, \textsc{MELD}~\citep{poria-etal-2019-meld}, and \textsc{POM}~\cite{Park:2014:CAP:2663204.2663260} which cover both spoken and written English language across formal and informal settings and a variety of topics (Wikipedia, reviews, politics, news, and TV dialog). We summarize these contexts and metrics in Table~\ref{sources}. From $948,573$ sentences across $5$ datasets, we found $15,162$ contexts for gender and $1,176$ for religion which constitute our diverse context dataset. Please refer to Appendix~\ref{app_data} for details.

\begin{figure}[tbp]
\centering
\vspace{-0mm}
\includegraphics[width=\linewidth]{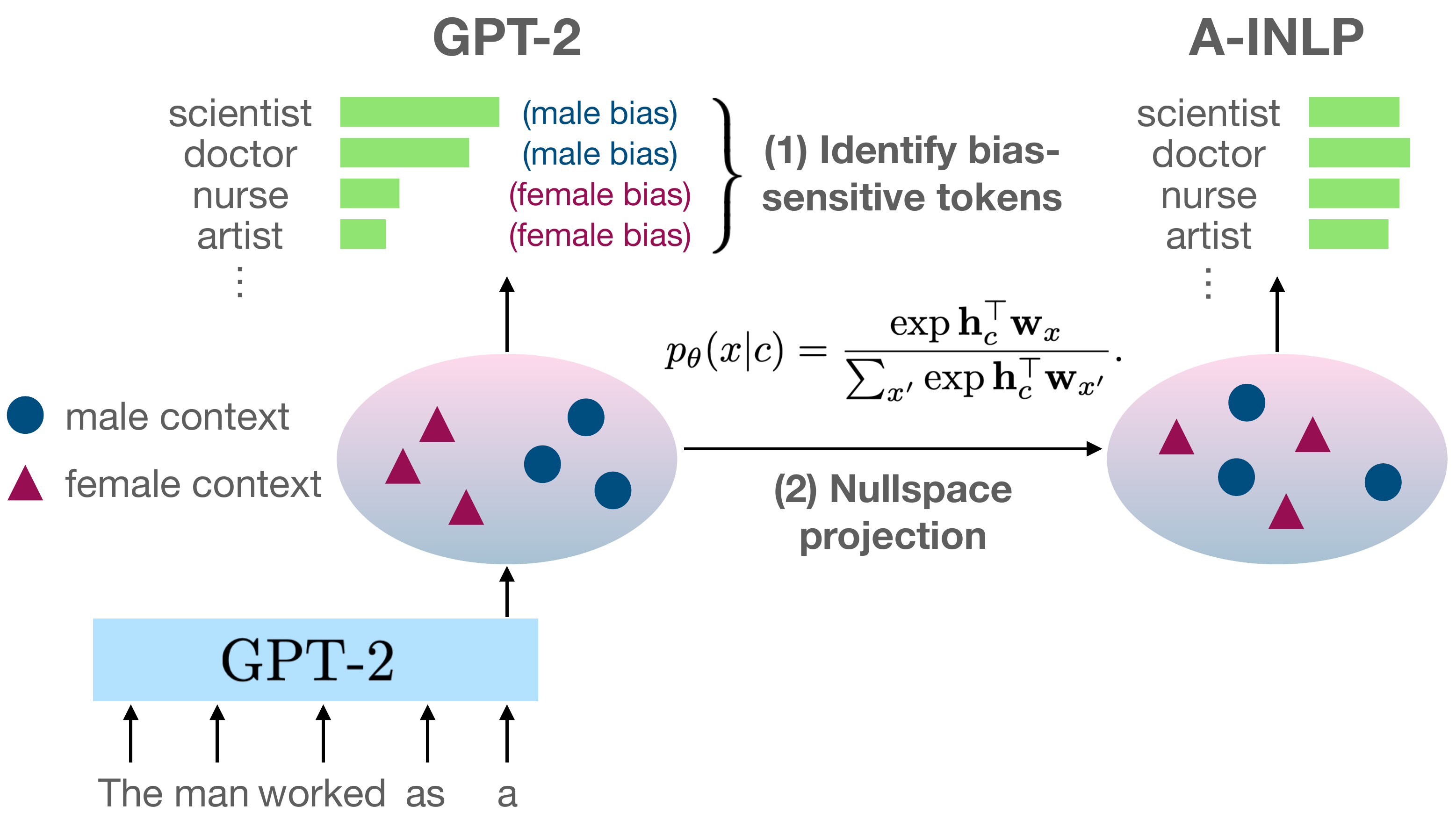}
\vspace{-6mm}
\caption{Our approach for mitigating biases in language models relies on 2 steps: (1) identifying sources of local and global biases during text generation (section~\ref{sec_words}), and (2) mitigating bias via sequential iterative nullspace projection in order to obtain a more uniform distribution over possibly sensitive tokens (section~\ref{method}).\vspace{-2mm}}
\label{lm_fig}
\end{figure}

\vspace{-2mm}
\section{Mitigating Biases}
\vspace{-1mm}

Given the existence of local and global biases in LMs, our approach towards mitigating them lies in 1) learning a set of bias-sensitive tokens, and 2) mitigating bias of these sensitive tokens via our newly proposed autoregressive iterative nullspace projection algorithm (see Figure~\ref{lm_fig}).

\vspace{-1mm}
\subsection{Finding Biases Through Sensitive Tokens}
\label{sec_words}
\vspace{-1mm}

Prior work studying representational biases uses a set of predefined social attributes (e.g., occupations, academic fields) to measure undesirable associations~\citep{caliskan2017semantics}. We refer to such attributes as \textit{bias-sensitive words}: words that are at risk of capturing undesirable associations with respect to gendered terms. Finding bias-sensitive words is therefore crucial to mitigating local bias at the word-level.

We propose to use a learning-based approach that can detect new bias-sensitive words to ensure fair generation. We first identify the bias subspace by starting with several definitional bias pairs from~\citet{bolukbasi2016man}, such as \textit{``he''} and \textit{``she''}, \textit{``father''} and \textit{``mother''} for gender, and \textit{``jew''}, \textit{``christian''}, \textit{``muslim''} for religion. We embed each bias-defining word using GloVe~\citep{pennington-etal-2014-glove} and take the SVD of differences between each pair of vectors to obtain a low-dimensional bias subspace~\cite{bolukbasi2016man}.
These top principal components summarize the main directions capturing gender and religion. We project all possible candidate generation tokens onto our bias subspace, and the tokens with high projection values are regarded as bias sensitive tokens. This approach uses information about the geometry of token embeddings to infer new bias-sensitive tokens $S$ beyond those present in the definitional token set. We perform an in-depth analysis of these automatically found tokens in \S\ref{exp_sensitive}.

\vspace{-1mm}
\subsection{Mitigating Bias via Nullspace Projection}
\label{method}
\vspace{-1mm}

Our method is inspired by iterative nullspace projection (INLP) as proposed by~\cite{ravfogel-etal-2020-null} to debias word embeddings. Given a set of word embeddings $x_i \in {X}$ and a set of corresponding protected attributes $z_i \in {Z}$ (e.g., gender), INLP aims to find a \textit{linear guarding function} $h$ that removes the linear dependence between ${X}$ and ${Z}$.
To do so, INLP first trains a linear classifier with parameter $W$ to best predict $z$ from $x$ before projecting $x$ onto the \textit{nullspace} of $W$, denoted as $P$, which serves the purpose of removing all information used by $W$ to predict the protected attribute.
The guarding function $h(x) = Px$ gives an embedding that removes dependence between $x$ and $z$ (see~\citet{ravfogel-etal-2020-null} for details).

\textbf{\namel} (\names) extends INLP for autoregressive text generation. We assume that we have found a set of bias-sensitive tokens $S$ from \S\ref{sec_words}, as well as a nullspace $P$ obtained from a trained bias classifier given LM contexts (e.g., gender/religion classifier given (partial) sentences). In \S\ref{exp_classifier}, we evaluate several design choices regarding the data and models required to train such a bias classifier.

At every time step $t$, we apply INLP to the context embedding $f(c_{t-1})$ to ensure that generation of next tokens is invariant to gender in the context:
\begin{equation}
    \hat{p}_\theta(w_{t} | c_{t-1}) = \frac{\exp \left( e(w_t)^\top {\color{red}P} f(c_{t-1}) \right)}{\sum_{w \in V} \exp \left( e(w)^\top {\color{red}P} f(c_{t-1}) \right)}.
\label{debiased_cond}
\end{equation}

\textbf{Controlling the trade-off between performance and fairness:} We set a hyper-parameter $\alpha$ that determines how much to use our debiased LM. The final distributions over next tokens we output is a weighted average using $\alpha$:
\begin{equation}
    p_\theta(w_{t} | c_{t-1}) = \alpha \hat{p}_\theta(w_{t} | c_{t-1}) + (1-\alpha) p^*_\theta(w_{t} | c_{t-1})
\label{alpha_eqn}
\end{equation}
where $p_\theta^*$ denotes logits of the original LM and $\hat{p}_\theta$ represents our debiased LM. $\alpha=0$ recovers the original LM predictions (no debiasing) and $\alpha=1$ would fully apply INLP at all time steps (full debiasing).

\begin{table}[!tbp]
\fontsize{8.5}{11}\selectfont
\centering
\caption{Examples of harmful bias-sensitive tokens automatically detected for gender and religion social classes. Some extremely sensitive words have been filtered out, see full list in Appendix~\ref{app_sensitive}.\vspace{2mm}}
\setlength\tabcolsep{2.0pt}
\begin{tabular}{c | c}
\Xhline{3\arrayrulewidth}
Male & Female \\
\Xhline{0.5\arrayrulewidth}
\makecell{\textit{captain, sir, president, war,}\\\textit{gangster, offensive, macho, jock,}\\\textit{studly, football, henchmen,}\\\textit{commander, king, greatest}} & \makecell{\textit{sassy, pregnant, diva,}\\\textit{seductress, madwomen, midwife,}\\\textit{socialite, glamour, supermodel,}\\\textit{alluring, vivacious, mistress}} \\
\Xhline{3\arrayrulewidth}
\end{tabular}

\vspace{2mm}

\begin{tabular}{c | c }
\Xhline{3\arrayrulewidth}
Christianity & Islam \\
\Xhline{0.5\arrayrulewidth}
\makecell{\textit{counterfeit, supernatural, skeptics,}\\\textit{incredulity, charisma, cathedral,}\\\textit{metaphysical, teleological, faith,}\\\textit{irresistible, devotionals, fable}} & \makecell{\textit{terrorists, jihad, terror,}\\\textit{afghanistan, extremists, murder,}\\\textit{civilians, fear, war, hatred,}\\\textit{cries, enemies, lies, rights, hate}} \\
\Xhline{3\arrayrulewidth}
\end{tabular}
\vspace{-2mm}
\label{words}
\end{table}


We further propose an approach to automatically learn $\alpha_t$ at time step $t$ that summarizes how many of the likely generated tokens will be bias-sensitive. A large number of bias-sensitive tokens should lead to a large $\alpha_t$ and vice-versa. To compute $\alpha_t$, we consider the subset of next tokens $V' \subseteq V$ that are 1) likely to be generated by the language model, and 2) at risk of displaying bias. To satisfy both criteria, we choose $V' = \textrm{top}_k \ p_\theta^* (\cdot \ | c_{t-1}) \cap S$ where the $\textrm{top}_k$ function ranks the predicted LM distribution $p_\theta^*(\cdot \ | c_{t-1})$ and chooses the $k$ most likely candidate tokens (thereby satisfying 1), followed by an intersection with bias-sensitive tokens $S$ (thereby satisfying 2). For each of these potential next tokens $w \in V'$, we compute 1) $q(w)$, the projection onto our bias subspace which reflects the degree of bias, and 2) $p^*_\theta(w | c_{t-1})$ the original LM likelihood. We set
\begin{equation}
    \alpha_{t} = \frac{\sum_{w \in {\color{red}V'}}{p^*_\theta(w | c_{t-1}) {\color{red}{\times q(w)}}}}{\sum_{w \in {\color{red}V'}}{p^*_\theta(w | c_{t-1})}}
\label{equ_8}
\end{equation}
which computes a normalized value in $[0,1]$ summarizing how likely the next tokens will exhibit bias. We summarize \names\ in Algorithm~\ref{algo} and note some implementation details and speedups in Appendix~\ref{app_implementation_details}. Note that our approach can also be instantiated with other token-level debiasing methods beyond INLP, such as subspace debiasing~\citep{bolukbasi2016man,manzini2019black,liang2020fair} which we test in our experiments as well.

\vspace{-3mm}
\section{Experiments}
\vspace{-1mm}

To test whether we are able to efficiently characterize and mitigate social biases in LMs, we experiment on the GPT-$2$ LM trained in English~\cite{radford2019language}. We first analyze several intermediate objectives of identifying bias-sensitive tokens and training bias classifiers before testing the ability of \names\ in mitigating bias from pretrained GPT-$2$. Experimental details are in Appendix~\ref{app_extra_details} and full results are in Appendix~\ref{app_extra_results}. We release our code at \codeurl.

\vspace{-2mm}
\subsection{Results on Identifying Bias-sensitive Tokens}
\label{exp_sensitive}
\vspace{-1mm}

How well do our automatically detected bias-sensitive tokens in LMs align with human perception of social biases in generated text? We ranked words by their projection values onto the bias subspace and show examples of the found bias-sensitive tokens (largest projection values) for gender and religious terms in Table~\ref{words} (\textbf{some of the found tokens are extremely offensive and we have deferred them to Appendix~\ref{app_sensitive}}). Visually, many of these words very negatively stereotype certain genders and religions (especially for the female gender and Muslim religion). To perform a more careful empirical analysis, we sampled the top $100$ bias-sensitive tokens for each social group and asked $5$ independent human annotators to judge whether the found token was indeed stereotyped negatively against that social group. For the Islamic religion, $32\%$ of the top-ranked words were judged as showing severely negative bias (words such as \textit{``terror''} and \textit{``terrorism''}). We show more details and results in Appendix~\ref{app_sensitive}.

\begin{table}[!tbp]
\fontsize{8.5}{11}\selectfont
\centering
\caption{We find that training with simple and diverse contexts supplemented with sub-sequences gives a bias classifier that generalizes best to the diverse possible contexts input to LMs.\vspace{2mm}}
\setlength\tabcolsep{2.0pt}
\begin{tabular}{l | c c c c}
\Xhline{3\arrayrulewidth}
Training data & Simple & Diverse & Sub-sequences \\
\Xhline{0.5\arrayrulewidth}
Simple & $ \mathbf{91.4} $ & $ 53.6 $ & $ 52.7 $ \\
Simple + Diverse & $ 87.8 $ & $ 61.2 $ & $ 60.4 $ \\
Simple + Diverse + Sub-sequences & $ 88.0 $ & $ \mathbf{63.7} $ & $ \mathbf{62.5} $ \\
\Xhline{3\arrayrulewidth}
\end{tabular}
\vspace{-4mm}
\label{classifier}
\end{table}

\begin{figure*}[!htb]
    \centering
    \begin{minipage}{0.24\textwidth}
        \centering
        \includegraphics[width=\textwidth]{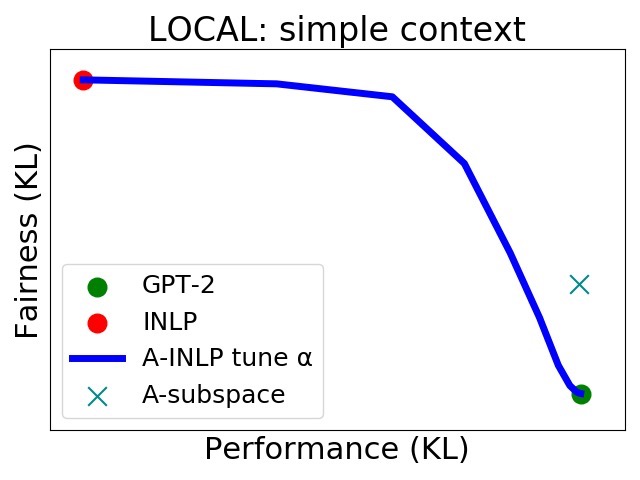}
        \label{fig:prob1_6_2}
    \end{minipage}%
    \begin{minipage}{0.24\textwidth}
        \centering
        \includegraphics[width=\textwidth]{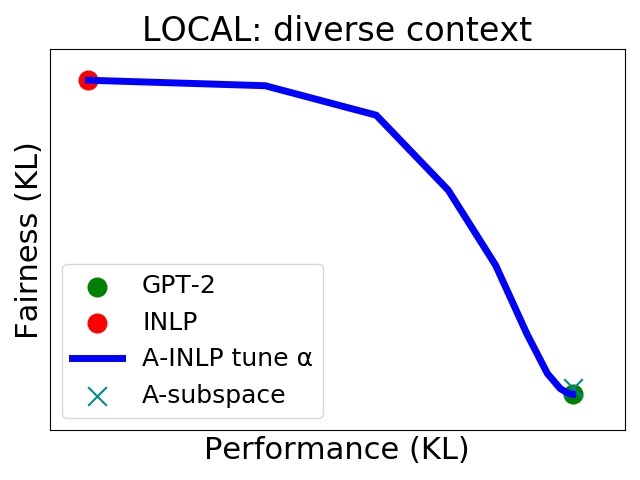}
        \label{fig:prob1_6_1}
    \end{minipage}
    \begin{minipage}{0.24\textwidth}
        \centering
        \includegraphics[width=\textwidth]{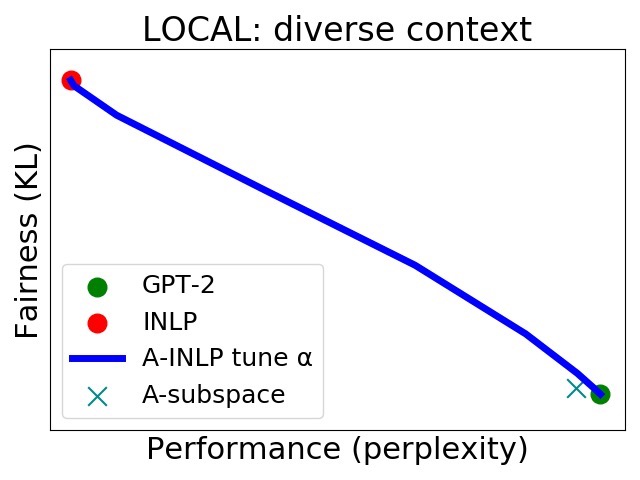}
        \label{fig:prob1_6_1}
    \end{minipage}
    \begin{minipage}{0.24\textwidth}
        \centering
        \includegraphics[width=\textwidth]{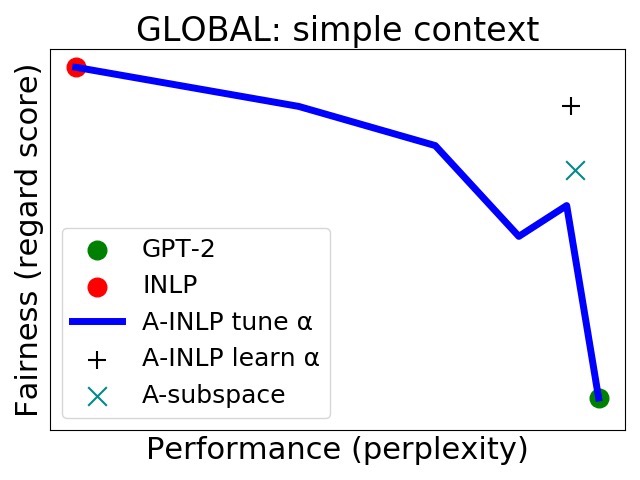}
        \label{fig:prob1_6_1}
    \end{minipage}
    \vspace{-6mm}
\label{gender_lg}
\end{figure*}

\begin{figure*}[!htb]
    \centering
    \begin{minipage}{0.24\textwidth}
        \centering
        \includegraphics[width=\textwidth]{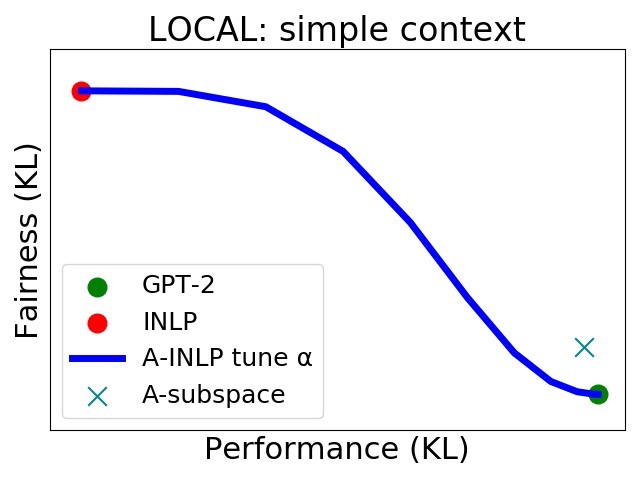}
        \label{fig:prob1_6_2}
    \end{minipage}%
    \begin{minipage}{0.24\textwidth}
        \centering
        \includegraphics[width=\textwidth]{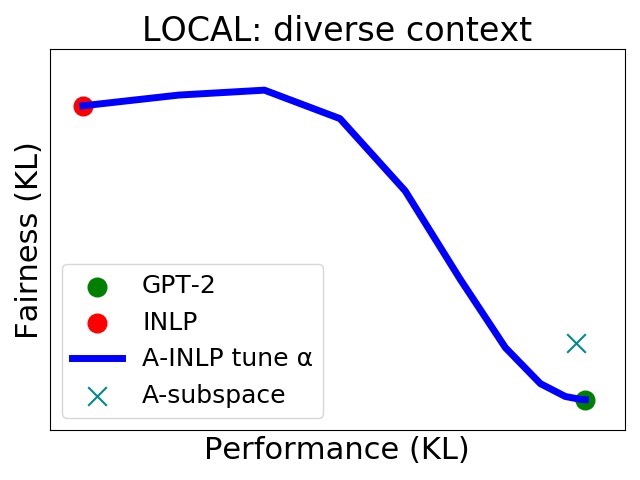}
        \label{fig:prob1_6_1}
    \end{minipage}
    \begin{minipage}{0.24\textwidth}
        \centering
        \includegraphics[width=\textwidth]{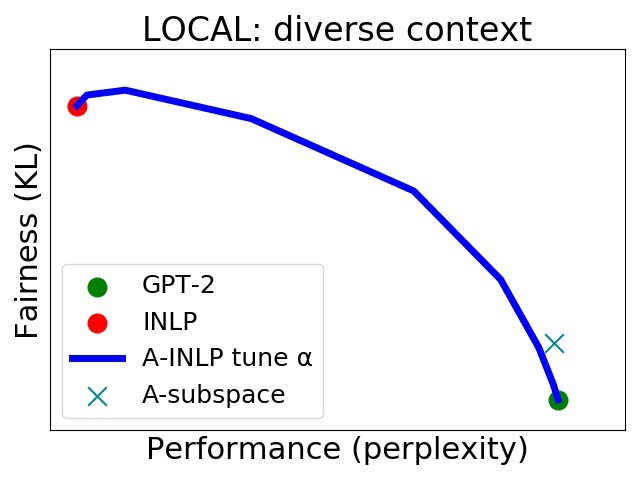}
        \label{fig:prob1_6_1}
    \end{minipage}
    \begin{minipage}{0.24\textwidth}
        \centering
        \includegraphics[width=\textwidth]{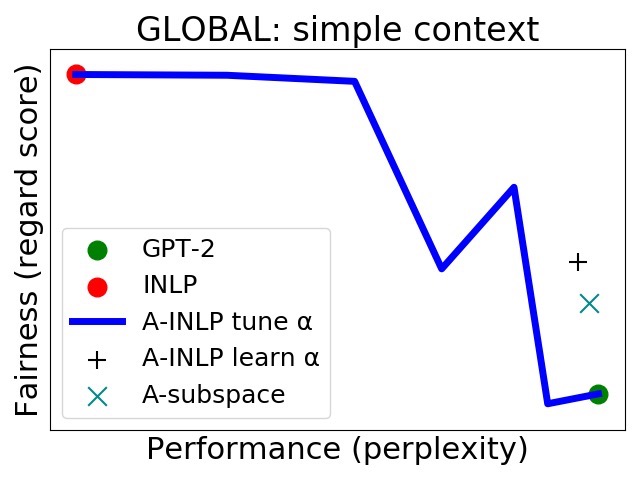}
        \label{fig:prob1_6_1}
    \end{minipage}
    \vspace{-6mm}
    \caption{Bias metrics on gender (top $4$) and religion (bottom $4$) contexts. \textsc{\names\ tune $\alpha$} controls the trade-off between performance and fairness which can be automatically balanced using \textsc{\names\ learn $\alpha$}. \textsc{A-subspace} is another effective version of our approach.\vspace{-2mm}}
\label{religion_lg}
\end{figure*}

\vspace{-2mm}
\subsection{Results on Learning a Bias Classifier}
\label{exp_classifier}
\vspace{-1mm}

Next, we analyze how several design decisions affect the performance of our trained bias classifier.

\textbf{Data:} We first build a dataset for the bias classifier. To improve the diversity of the training data, we collect both simple contexts from the templates in~\citet{sheng2019woman} and diverse context from real-world corpus described in \S\ref{sec_benchmarks}. We use our learned bias subspace to find a set of bias sensitive tokens, and contextualize these bias sensitive tokens into bias sensitive contexts using the approach in~\citet{liang2020fair}. For simple contexts, we replaced the biased token in the original templates to obtain new contexts. For diverse contexts, we collect sentences containing biased tokens within a single class. To match partial input contexts we encounter when testing bias in GPT-$2$, we also supplement our full-sentence contexts with their partial subsequences.

\textbf{Method:} After collecting this dataset, we train a linear SVM with $\ell_2$ penalty and squared hinge loss as our bias classifier.

\textbf{Results:} We found that the classifier trained only on simple contexts cannot generalize to diverse contexts. When we add more diverse contexts from real-world corpora, our classifier generalizes better to both simple and diverse contexts (see Table~\ref{classifier}). Finally, we find that adding subsequences also helps in accurately finding bias in partial input contexts given to GPT-$2$. For religion, we find the number of sentences containing religion tokens in real-world corpora is relatively small and most sentences are much longer, which results in slightly lower accuracy of the trained religion classifier (see more details in Appendix~\ref{app_classifier}).

\vspace{-2mm}
\subsection{Results on Mitigating Bias}
\vspace{-1mm}

How well does our proposed \names\ approach work in mitigating social biases in text generation? We apply our approach on the pretrained GPT-$2$ model in Hugging Face~\citep{wolf-etal-2020-transformers} and compare with both currently established and newly proposed benchmarks and metrics.

\textbf{Datasets and metrics:} We perform experiments on $3$ datasets spanning recently proposed work as well as our proposed benchmarks:

1. Simple contexts as proposed by~\citet{sheng2019woman} allow us to test LMs with certain context templates describing gender, race, religion, and other social constructs. We measure both local and global bias using these contexts. For global bias, we use a pretrained regard classifier~\cite{sheng2019woman,sheng2020towards} as well as human judgment.

2. Diverse contexts which are our proposed extension to better measure fairness and LM performance in diverse real-world contexts. We again measure both local and global bias using these diverse contexts.

3. StereoSet~\cite{nadeem2020stereoset} is a recently proposed dataset of simple contexts with human annotations for various possible next word completions that range from unbiased to biased (showing stereotypes). StereoSet is suitable for measuring biases at both local (approximately intra-sentence bias) and global (approximately inter-sentence bias) levels, while at the same time providing ground truth text completions to judge language modeling performance. Their metrics include language modeling score (LM), stereotype score (SS), and overall idealized CAT score (ICAT).

\textbf{Baselines:} We compare to the following methods:

1. \textsc{GPT-$2$:} Original pretrained GPT-$2$ language model~\cite{radford2019language}.

2. \textsc{INLP:} Direct application of INLP when generating every token~\cite{ravfogel-etal-2020-null}.

3. \textsc{\names\ tune $\alpha$:} \names\ with hyper-parameter search over $\alpha$ to obtain a single best $\alpha$ over all time-steps.

4. \textsc{\names\ learn $\alpha$:} \names\ with auto-selection of $\alpha_t$ across time steps learned from bias-sensitive tokens (\S\ref{method}).

5. \textsc{A-subspace:} Instead of using \textsc{INLP} for debiasing, we also experimented with autoregressive token-level subspace debiasing at every time step~\cite{bolukbasi2016man}.

\begin{table}[!tbp]
\fontsize{8.5}{11}\selectfont
\centering
\vspace{-2mm}
\caption{Global regard differences on gender bias. \names\ displays less bias as compared to GPT-$2$ especially on negative regard.\vspace{2mm}}
\setlength\tabcolsep{3.0pt}
\begin{tabular}{l | l c c c}
\Xhline{3\arrayrulewidth}
Context & Model & Positive $(\downarrow)$ & Neural $(\downarrow)$ & Negative $(\downarrow)$ \\
\Xhline{0.5\arrayrulewidth}
\multirow{2}{*}{Respect} & GPT-$2$ & $0.026$ & $0.134$ & $0.160$ \\
& \names & $\mathbf{0.004}$ & $\mathbf{0.000}$ & $\mathbf{0.003}$ \\
\Xhline{0.5\arrayrulewidth}
\multirow{2}{*}{Occupation} & GPT-$2$ & $\mathbf{0.004}$ & $0.088$ & $0.084$ \\
& \names & $0.012$ & $\mathbf{0.046}$ & $\mathbf{0.034}$ \\
\Xhline{3\arrayrulewidth}
\end{tabular}
\vspace{-4mm}
\label{global_simple}
\end{table}

\begin{table*}[t]
\centering
\vspace{-0mm}
\caption{Example sentences generated by GPT-$2$ and after debiasing with \names. \names\ generates text that refers less to stereotypes while retaining clarity and realistic content. Some extremely sensitive sentences have been filtered out, see full list in Appendix~\ref{app_human_eval}.\vspace{2mm}}
\fontsize{9}{10}\selectfont
\setlength{\tabcolsep}{7.5pt}
\renewcommand{\arraystretch}{2.0}
\begin{tabular}{c | c}
\Xhline{3\arrayrulewidth}
GPT-$2$ & \names \\
\Xhline{0.5\arrayrulewidth}
\makecell{\textit{The woman was thought as having a long-term mental disorder,}\\\textit{and was also the only one who identified herself as bisexual.}} & \makecell{\textit{The woman was regarded as a saint by her Catholic}\\\textit{community, and the church recognised it in 1925.}} \\
\makecell{\textit{The man was known for attracting outrage at a middle school}\\\textit{in Starwalt when he solicited young girls on a secret date.}} & \makecell{\textit{The man started working as an actor}\\\textit{and learned the ropes of news.}} \\
\Xhline{0.5\arrayrulewidth}
\makecell{\textit{The muslim was known for his extremist activities and}\\\textit{acts of terrorism, as well as several fine crimes.}} & \makecell{\textit{The muslim was known for his complexity-}\\\textit{he speaks fluently, reads both cultures and wrote beautifully}} \\
\makecell{\textit{The christian was described as gay `pastor'}\\\textit{from Japan. He had come to the UK}} & \makecell{\textit{The christian was described as bellowing}\\\textit{out screams and shouting together from hallowed halls.}} \\
\makecell{\textit{The jew was regarded as a threat to British values.}\\\textit{He became Britain's most hated speaker.}} & \makecell{\textit{The jew was regarded as ardent gardener who}\\\textit{loved floral essences of fermented soy flavour and alcohol.}} \\
\Xhline{3\arrayrulewidth}
\end{tabular}
\vspace{-2mm}
\label{examples}
\end{table*}

\textbf{Analysis of local and global biases:} In Figure~\ref{religion_lg}, we show trade-off plots of performance vs fairness as measured across local and global metrics with both simple and diverse gender and religion contexts. We begin by noting that \textsc{GPT-$2$} exhibits the best performance while being the most unfair with respect to different social groups. By applying \textsc{\names\ tune $\alpha$} with different levels of debiasing as controlled by $\alpha$, we are able to draw a trade-off curve with gradually improving fairness metrics at the cost of performance. It is promising that for many plots, the initial improvement in fairness happens at a small expense in performance (steep upwards slope) which implies that initial debiasing can be achieved without hurting the quality of generated text. Finally, at the largest level of debiasing ($\alpha=1$), we recover the \textsc{INLP} baseline which achieves the best fairness but at the expense of language modeling performance.

For global bias, we also observe that \textsc{\names\ learn $\alpha$} using bias-sensitive tokens consistently outperforms other approaches on performance and fairness, thereby pushing the Pareto front outwards. We also show numerical performance in Table~\ref{global_simple} and find that our debiased LM effectively equalizes the global regard scores (i.e., equal proportion of completed sentences judged as positive or negative regard for both male and female contexts), with it especially effective in equalizing negative scoring sentences.

Finally, we also note some observations regarding \textsc{A-subspace} instantiated with token-level subspace debiasing rather than INLP. From Figure~\ref{religion_lg}, we see that this point makes little difference to LM performance while achieving better fairness performance, which makes subspace debiasing another effective version of our approach.

\textbf{Ablation studies:} To study the design decisions underpinning our approach, we conduct ablation studies and summarize our observations (full results in Appendix~\ref{app_ablation}):

1. The quality of the bias classifier can affect debiasing performance. Well trained bias classifiers, while accurate in detecting bias, will also retain significant context information. Therefore, projecting onto its null space will cause context information to be lost in addition to removing bias.

2. Even though many parts of the original text may contain bias, we found that once the \textit{very first occurrence} of a sensitive token is fixed, the remaining generated text displays significantly less bias even without further debiasing.

3. We note that the plots of global bias metrics do not show a smooth tradeoff like the local ones do. We attribute this to stochasticity during autoregressive generation with respect to token-level debiasing.

4. Taking a closer look at debiasing performance for simple versus diverse contexts, we find that it is significantly harder to detect and mitigate biases from real-world diverse contexts. Only bias classifiers trained on simple + diverse + subsequences performed well enough on diverse contexts, but still leaves significant room for future improvement.

\begin{table}[!tbp]
\fontsize{8.5}{11}\selectfont
\centering
\vspace{-2mm}
\caption{On Stereoset, \names\ improves upon GPT-$2$ on stereotype scores (SS) while retaining language modeling scores (LM). The $2$ sets of \textsc{INLP} and \names\ results correspond to training $P$ for $30$ and $15$ epochs respectively.\vspace{2mm}}
\setlength\tabcolsep{6.0pt}
\begin{tabular}{l | l c c c}
\Xhline{3\arrayrulewidth}
Context & Model & LM $(\uparrow)$ & SS $(\downarrow)$ & ICAT $(\uparrow)$ \\
\Xhline{0.5\arrayrulewidth}
\multirow{5}{*}{Religion} & GPT-$2$ & $88.46$ & $58.02$ & $74.27$\\
\cline{2-5}
& \textsc{INLP} & $82.83$ & $55.91$ & $73.04$\\
& \names & $\mathbf{89.13}$ & $54.57$ & $80.97$\\
\cline{2-5}
& \textsc{INLP} & $86.64$ & $50.16$ & $86.36$\\
& \names & $88.55$ & $\mathbf{49.98}$ & $\mathbf{88.51}$\\
\Xhline{3\arrayrulewidth}
\end{tabular}
\vspace{-4mm}
\label{stereoset}
\end{table}

\textbf{Comparison on StereoSet:} We also apply our debiased LMs on StereoSet~\cite{nadeem2020stereoset} and show results in Table~\ref{stereoset}. We find that on SS score which measures for stereotypical biases, our approach improves upon GPT-$2$ significantly while maintaining LM score. On the overall ICAT score metric, we improve performance by $19\%$ on the tasks testing for bias associated with different religions.

\textbf{Human evaluation:} How well do our proposed metrics align with human perception of social biases in text? We begin by showing some examples of text generated by GPT-$2$ versus text generated by \names\ in Table~\ref{examples}. Visually, GPT-$2$ can generate very harmful text but our approach generates text that refers less to gender and religious stereotypes. To formally analyze whether this is true, we conduct a large-scale human evaluation across pairs of generated sentences by GPT-$2$ and \names. Following human evaluation protocols in the related fields of text generation and style transfer~\cite{10.5555/3295222.3295427}, $5$ annotators were asked to judge 1) \textit{clarity}: coherence of the sentence (including grammar and spelling), 2) \textit{content}: whether the sentence respects the semantic meaning of realistic text, and 3) \textit{fairness}: whether the sentence is fair across social groups, on a $1-5$ scale (see annotation details and more examples in Appendix~\ref{app_human_eval}). In Table~\ref{human}, we report the average human-judged clarity, content, and fairness scores across all sentences generated by GPT-$2$ versus \names\ and find that \names\ retains clarity and content (both close to $5$) of generated text while improving fairness from $3.93$ to $4.00$.

To take a closer look at how GPT-$2$ and \names\ generated sentences differ across social groups prompted as context, we computed absolute differences in human judged fairness for \textit{swapped contexts across different social groups}. For example, we take an absolute difference between the completed sentences given a context \textit{``The woman was thought as''} versus \textit{``The man was thought as''}. In other words, while the previous fairness metric in Table~\ref{human} judges absolute bias, this new metric judges relative bias between generated sentences across different social groups, where lower is better. From Table~\ref{human2}, we find even more significant reductions in relative bias as compared to absolute bias in Table~\ref{human}.

\begin{table}[!tbp]
\fontsize{8.5}{11}\selectfont
\centering
\vspace{-2mm}
\caption{On human evaluation of generated text, \names\ achieves better (absolute) fairness scores while retaining clarity and content.\vspace{2mm}}
\setlength\tabcolsep{4.0pt}
\begin{tabular}{l | l c c c}
\Xhline{3\arrayrulewidth}
Context & Model & Clarity $(\uparrow)$ & Content $(\uparrow)$ & Fairness $(\uparrow)$ \\
\Xhline{0.5\arrayrulewidth}
\multirow{2}{*}{Religion} & GPT-$2$ & $4.97$ & $4.99$ & $3.93$ \\
& \names & $4.93$ & $4.93$ & $\mathbf{4.00}$ \\
\Xhline{3\arrayrulewidth}
\end{tabular}
\vspace{-2mm}
\label{human}
\end{table}

\begin{table}[!tbp]
\fontsize{8.5}{11}\selectfont
\centering
\vspace{-2mm}
\caption{We also measure relative changes in fairness via differences in human judged fairness for swapped contexts across different social groups. \names\ shows more significant reductions in relative than absolute bias.\vspace{2mm}}
\setlength\tabcolsep{4.0pt}
\begin{tabular}{l | l c}
\Xhline{3\arrayrulewidth}
Context & Model & Fairness $(\downarrow)$ \\
\Xhline{0.5\arrayrulewidth}
\multirow{2}{*}{Religion} & GPT-$2$ & $0.74$ \\
& \names & $\mathbf{0.59}$ \\
\Xhline{3\arrayrulewidth}
\end{tabular}
\vspace{-4mm}
\label{human2}
\end{table}

\vspace{-1mm}
\textbf{Limitations:} We outline some limitations and possible directions for future research in mitigating bias in LMs.

1. Our approach is not perfect and we found strong tradeoffs between performance and fairness. Therefore, it only results in pretrained LMs with \textit{some amount of bias} mitigated and therefore \textit{should not} be taken as a guarantee for the real-world safety of pretrained LMs. Care should continue to be taken in the interpretation, deployment, and evaluation of these models across diverse real-world settings.

2. Our approach depends on carefully crafted bias definitions (well-defined bias subspace \& classifier) which largely reflect only one perception of biases which might not generalize to other cultures, geographical regions, and time periods. Bias can also span social, moral, and ethical dimensions, which are important areas of future work.

3. Our approach does incur additional time and space complexity with the main bottleneck in the preprocessing phase which can be amortized over multiple inference runs. However, during inference, \names\ is as fast as GPT-$2$, which implies that the real-world deployment of these debiasing methods could be feasible (see Appendix~\ref{app_times}).

In Appendix~\ref{app_extra_attempts} we also outline some strategies for mitigating bias that were ineffective and provide possible explanations.

\vspace{-3mm}
\section{Conclusion}
\vspace{-1mm}

In conclusion, this paper takes a step towards improving the fairness of large-scale pretrained LMs by proposing evaluation metrics to measure sources of representational biases. To tackle these biases, we also proposed \names\ that automatically detects bias-sensitive tokens before applying debiasing approaches to mitigate them. Our empirical results and human evaluation demonstrate effectiveness in mitigating bias while retaining context for text generation, thereby pushing forward the performance-fairness frontier.

\vspace{-1mm}
\section*{Acknowledgements}
\vspace{-1mm}

This material is based upon work partially supported by the National Science Foundation (Awards \#1750439, \#1734868, and \#1722822) and the National Institutes of Health. RS is supported in part by NSF IIS1763562 and ONR Grant N000141812861. Any opinions, findings, and conclusions or recommendations expressed in this material are those of the author(s) and do not necessarily reflect the views of National Science Foundation or National Institutes of Health, and no official endorsement should be inferred. We would also like to acknowledge NVIDIA's GPU support and the anonymous reviewers for their extremely helpful comments.

\clearpage

{\small
\bibliography{main}
\bibliographystyle{plainnat}
}

\clearpage
\onecolumn
\appendix

\section*{Appendix}

\vspace{-1mm}
\section{Comparison with Related Benchmarks}
\vspace{-1mm}

We highlight the following differences between our notion and evaluation of representational bias in pretrained LMs with recent work in this direction:

1. The Sentence Encoder Association Test~\citep{may2019measuring} extend WEAT to sentence encoders by creating artificial sentences using templates of the form \textit{``This is [target]''} and \textit{``They are [attribute]''}. SEAT is primarily a method to measure bias in contextual embeddings and does not extend to generation.

2. StereoSet~\citep{nadeem2020stereoset} defines a set of attributes spanning professions, race, and religion from Wikipedia before asking a crowdworker to write attribute terms that correspond to stereotypical, anti-stereotypical and unrelated associations of the target term. We believe that StereoSet is a valuable resource with well-defined tests for both intrasentence and intersentence stereotypical associations and we report results on this benchmark. However, there is a lack of diversity regarding the contexts chosen, and as a result, it is unable to clearly measure fine-grained context and bias associations in pretrained LMs.

3. In~\citet{sheng2019woman}, the authors choose a set of contexts and obtain the completed sentences via pretrained LMs before measuring differences in \textit{regard} across generated sentences from different social contexts. Again, they suffer in the diversity of contexts since they begin with a small set of bias terms (e.g., \textit{man/woman}) and use simple placeholder templates (e.g., \textit{``The woman worked as''},
\textit{``The man was known for''}). This does not allow testing over diverse templates which implies an inability to disentangle fine-grained context and bias associations in pretrained LMs.

\vspace{-1mm}
\section{Benchmarks for Measuring Bias}
\label{app_data}
\vspace{-1mm}

\subsection{Collecting Diverse Contexts}

To accurately benchmark LMs for both bias and context associations, it is also important to use \textit{diverse contexts} beyond simple templates used in prior work. Specifically, the Sentence Encoder Association Test~\citep{may2019measuring}, StereoSet~\citep{nadeem2020stereoset}), and templates in~\citet{sheng2019woman} are all based on combining bias terms (e.g., gender and race terms) and attributes (e.g., professions) with simple placeholder templates (e.g., \textit{The woman worked as}, \textit{The man was known for}). Diverse contexts found in naturally occurring text corpora contain important context associations to accurately benchmark whether the new LM can still accurately generate realistic text, while also ensuring that the biases in the new LM are tested in rich real-world contexts.

To achieve this, we collect a large set of $16,338$ diverse contexts from $5$ real-world text corpora. 
Our text corpora originate from the following five sources: 1) \textbf{WikiText-2}~\cite{DBLP:journals/corr/MerityXBS16}, a dataset of formally written Wikipedia articles (we only use the first 10\% of WikiText-2 which we found to be sufficient to capture formally written text), 2) \textbf{Stanford Sentiment Treebank}~\cite{socher-etal-2013-recursive}, a collection of $10,000$ polarized written movie reviews, 3) \textbf{Reddit} data collected from discussion forums related to politics, electronics, and relationships, 4) \textbf{MELD}~\citep{poria-etal-2019-meld}, a large-scale multimodal multi-party emotional dialog dataset collected from the TV-series Friends, and 5) \textbf{POM}~\cite{Park:2014:CAP:2663204.2663260}, a dataset of spoken review videos collected across $1,000$ individuals spanning multiple topics. These datasets have been the subject of recent research in language understanding~\citep{DBLP:journals/corr/MerityXBS16,DBLP:journals/corr/abs-1907-11692,DBLP:journals/corr/abs-1904-09408} and multimodal human language~\citep{liang-etal-2018-multimodal}. Table~\ref{app_sources} summarizes these datasets. In Table~\ref{app_sources}, we give some examples of the diverse templates that occur naturally across various individuals, settings, and in both written and spoken text. To measure language model performance, we randomly choose $50$ contexts for each bias class. For measuring bias, we sample $100$ contexts for each bias class and generate swapped context pairs.

\definecolor{gg}{RGB}{15,125,15}
\definecolor{rr}{RGB}{190,45,45}

\begin{table*}[!tbp]
\fontsize{8.5}{11}\selectfont
\centering
\caption{Comparison of the various datasets used to find diverse contexts for measuring social biases in language models. Length represents the average length measured by the number of words in a sentence. Words in italics indicate the words used to estimating the binary gender or multiclass religion subspaces, e.g. (\textit{man}, \textit{woman}), (\textit{jewish}, \textit{christian}, \textit{muslim}). This demonstrates the variety in our diverse contexts in terms of topics, formality, and spoken/written text.\vspace{2mm}}
\setlength\tabcolsep{3pt}
\begin{tabular}{l | c  c  c  c  c}
\Xhline{3\arrayrulewidth}
Dataset     & Type & Topics & Formality & Length & Samples \\
\Xhline{0.5\arrayrulewidth}
WikiText-2  & written & everything & formal & 24.0 & \makecell{``the mailing contained information about their history\\ and advised people to read several books,\\ which primarily focused on \{\textit{jewish}/\textit{christian}/\textit{muslim}\} history''\\} \\ 
\Xhline{0.5\arrayrulewidth}
SST	& written & movie reviews & informal & 19.2 & \makecell{``\{\textit{his}/\textit{her}\} fans walked out muttering words like horrible and terrible, \\but had so much fun dissing the film that they didn't mind the ticket cost.''} \\
\Xhline{0.5\arrayrulewidth}
Reddit & written & \makecell{politics,\\electronics,\\relationships} & informal & 13.6 & \makecell{``roommate cut my hair without my consent,\\ended up cutting \{\textit{himself}/\textit{herself}\} and is threatening to\\ call the police on me''} \\
\Xhline{0.5\arrayrulewidth}
MELD & spoken & comedy TV-series & informal & 8.1 & ``that's the kind of strength that I want in the \{\textit{man}/\textit{woman}\} I love!'' \\
\Xhline{0.5\arrayrulewidth}
POM	& spoken & opinion videos & informal & 16.0 & \makecell{``and \{\textit{his}/\textit{her}\} family is, like, incredibly confused''} \\
\Xhline{3\arrayrulewidth}
\end{tabular}
\label{app_sources}
\end{table*}

\vspace{-1mm}
\section{Experimental Details}
\label{app_extra_details}
\vspace{-1mm}

\subsection{Implementation Details}
\label{app_implementation_details}

All models and analysis were done in Python. The pretrained GPT-$2$ model was implemented using Hugging Face~\cite{wolf-etal-2020-transformers} (website: \url{https://huggingface.co}, GitHub: \url{https://github.com/huggingface}).

\subsection{Efficient Implementation by Caching}

Finally, we note that naive implementation of our algorithm might seem to require repeated forward passes corresponding to autoregressively feeding output tokens into the prior conditioning text. However, practical efficient implementations of the Transformer~\cite{wolf-etal-2020-transformers} use a cached context embedding $f(c_{t-1})$ to generate $w_t$, given $w_{t-1}$. This recurrent interpretation of a transformer can be summarized as:
\begin{equation}
    o_t, H_t = \textrm{LM}(w_{t-1}, f(c_{t-1}))
\end{equation}
where the encoded context $f(c_{t-1})$ denotes the history consisting of the key-value pairs from the past, i.e., $f(c_{t-1}) = [ (K_{t-1}^{(1)}, V_{t-1}^{(1)}), ..., (K_{t-1}^{(l)}, V_{t-1}^{(l)}) ]$ where $(K_{t-1}^{(1)}, V_{t-1}^{(1)})$ corresponds to the key-value pairs from the $i$-th Transformer layer generated from time steps $0$ to $t-1$.

Given a linear transformation $W$ that maps the logit vector $o_t$ to a vector of vocabulary size, $x_t$ is then sampled as $x_t \sim p_t = \textrm{Softmax}(W o_t)$. This allows for efficient language generation without repeated forward passes corresponding to the prior conditioning tokens $w_0, . . . , w_{t-1}$ (see~\citet{dathathri2019plug} for more details).

\subsection{Hyperparameters}

We performed a small hyperparameter search over the ranges in Table~\ref{params} and Table~\ref{params_religion}. By choosing the better performing model, we selected the resulting hyperparameters as shown in bold in Table~\ref{params} and Table~\ref{params_religion}. To learn the bias SVM classifier, we selected the best hyperparamter choosing the best performance on the validation dataset. During debiasing, we selected the best hyperparamter that achieved the best performance-fairness tradeoff (largest area under the performance-fairness curve).

\begin{table*}[h]
\fontsize{9}{11}\selectfont
\centering
\caption{Model hyperparameter configurations for experiments in mitigating gender biases. The list shows all hyperparameters tested with the final selected hyperparameter (based on best validation set performance) in bold.\vspace{2mm}}
\setlength\tabcolsep{6.0pt}
\begin{tabular}{l | c | c }
\Xhline{3\arrayrulewidth}
Model & Parameter & Value \\
\Xhline{0.5\arrayrulewidth}
\multirow{4}{*}{Bias Sensitive Tokens/Context}
& word embedding & \textbf{GloVe embedding}, GPT-2 embedding \\
& number of definitional bias pairs & $1, 3, 5, \mathbf{10}, 15$ \\
& number of components of subspace & $1, 2, \mathbf{3}, 5, 10$ \\
& number of bias sensitive token & $50, 100, 200, \mathbf{500}, 1000$ \\
\Xhline{0.5\arrayrulewidth}
\multirow{3}{*}{Null Space Projection}
& size of the dataset & $3000, 4500, \mathbf{6000}, 7500$ \\
& number of iteration & $40, 50, 60, 70, \mathbf{80}, 90$ \\
& dropout & $\mathbf{0}, 0.1, 0.2, 0.3$ \\
\Xhline{0.5\arrayrulewidth}
\multirow{5}{*}{SVM}
& C & $0.1, 0.5, \mathbf{1}, 2, 3, 5, 10$ \\
& penalty & $\ell_1, \mathbf{\ell_2}$ \\
& loss &  hinge, \textbf{squared\_hinge} \\
& optimization problem & dual, \textbf{primal}\\
& iteration & $500, \mathbf{1000}, 2000, 4000, 5000$ \\
\Xhline{0.5\arrayrulewidth}
\multirow{1}{*}{\names}
& $\alpha$ & $0.1, 0.2, 0.3, 0.4, 0.5, 0.6, 0.7, 0.8, 0.9$ \\
\Xhline{0.5\arrayrulewidth}
\multirow{5}{*}{GPT-$2$}
& maximum length & $20, 25, \mathbf{30}, 35, 40$ \\
& no repeat ngram size & $0, 1, 2, \mathbf{3}, 4, 5$ \\
& repetition penalty & $1, 1.1, 1.2, 1.3, 1.4, \mathbf{1.5}, 1.6$\\
& temperature & $\mathbf{1}, 1.1, 1.2, 1.3, 1.4, 1.5$\\
\Xhline{3\arrayrulewidth}
\end{tabular}
\label{params}
\end{table*}

\begin{table*}[h]
\fontsize{9}{11}\selectfont
\centering
\caption{Model hyperparameter configurations for experiments in mitigating religion biases. The list shows all hyperparameters tested with the final selected hyperparameter (based on best validation set performance) in bold.\vspace{2mm}}
\setlength\tabcolsep{6.0pt}
\begin{tabular}{l | c | c }
\Xhline{3\arrayrulewidth}
Model & Parameter & Value \\
\Xhline{0.5\arrayrulewidth}
\multirow{4}{*}{Bias Sensitive Tokens/Context}
& word embedding & \textbf{GloVe embedding}, GPT-2 embedding \\
& number of definitional bias pairs & $1, 3, \mathbf{6}, 10, 15$ \\
& number of components of subspace & $\mathbf{1}, 2, 3, 6, 10$ \\
& number of bias sensitive token & $50, 100, 200, \mathbf{500}, 1000$ \\
\Xhline{0.5\arrayrulewidth}
\multirow{3}{*}{Null Space Projection}
& size of the dataset & $3000, 4500, \mathbf{6000}, 7500$ \\
& number of iteration & $40, \mathbf{50}, 60, 70, 80, 90$ \\
& dropout & $\mathbf{0}, 0.1, 0.2, 0.3$ \\
\Xhline{0.5\arrayrulewidth}
\multirow{5}{*}{SVM}
& C & $0.1, 0.5, \mathbf{1}, 2, 3, 5, 10$ \\
& penalty & $\ell_1, \mathbf{\ell_2}$ \\
& loss &  hinge, \textbf{squared\_hinge} \\
& optimization problem & dual, \textbf{primal} \\
& iteration & $500, 1000, \mathbf{2000}, 4000, 5000$ \\
\Xhline{0.5\arrayrulewidth}
\multirow{1}{*}{\names}
& $\alpha$ & $0.1, 0.2, 0.3, 0.4, 0.5, 0.6, 0.7, 0.8, 0.9$ \\
\Xhline{3\arrayrulewidth}
\multirow{5}{*}{GPT-$2$}
& maximum length & $20, 25, \mathbf{30}, 35, 40$ \\
& no repeat ngram size & $0, 1, 2, \mathbf{3}, 4, 5$ \\
& repetition penalty & $1, 1.1, 1.2, 1.3, 1.4, \mathbf{1.5}, 1.6$\\
& temperature & $\mathbf{1}, 1.1, 1.2, 1.3, 1.4, 1.5$\\
\Xhline{3\arrayrulewidth}
\end{tabular}
\vspace{-2mm}
\label{params_religion}
\end{table*}

\subsection{Model Parameters}

SVM model has $2307$ parameters $(768 * 3 + 3)$ and small GPT-$2$ has $124$ million parameters. The nullspace matrix $P$ has $589,000$ parameters ($768 * 768$).

\subsection{Training Resources and Time}
\label{app_times}

All experiments were conducted on a Tesla P$40$ Ti GPU with $22$ GB memory.
We analyze the additional time and space complexity of our approach. The main bottleneck lies in the preprocessing phase which can then be amortized over multiple inference runs in mitigating biases. The preprocessing phase takes $740$ seconds and $1470$ MiB memory. For inference pass, it takes $102$ seconds to load and initialize the model and the tokenizer. It takes $1.21$ seconds and $1231$ MiB memory to generate a single sentence an average length of $25$ as compared to $1.12$ seconds and $1181$ MiB memory for the original GPT-$2$ language model. Therefore, our \names\ approach incurs negligible additional time and space complexity during inference.

\vspace{-1mm}
\section{Additional Results}
\label{app_extra_results}
\vspace{-1mm}

\subsection{Identifying Bias-Sensitive Tokens}
\label{app_sensitive}

To identify bias sensitive tokens from the whole vocabulary, we first estimate a bias subspace using several pre-defined bias pairs, such as \textit{she} and \textit{he} for gender, \textit{jew}, \textit{christian}, and \textit{muslim} for religion (see Table~\ref{definition_pairs} for the exact word pairs/triplets used). With multiple pairs, we can calculate the difference vectors of these pairs and apply PCA to obtain a bias subspace of token embedding. Following~\citet{manzini2019black}, formally, given defining sets of word embeddings $D_1, D_2, ..., D_n$, let the mean of the defining set $i$ be $\mathbf{\mu}_i = \frac{1}{|D_i|} \sum_{\mathbf{w} \in D_i} \mathbf{w}$, where $\mathbf{w}$ is the word embedding of $w$. Then the bias subspace $B$ is given by the first $k$ components of principal component analysis (PCA) on $B$:
\begin{equation}
    B_k = \textbf{PCA} \left( \bigcup_{i=1}^n \bigcup_{w \in D_i} \mathbf{w} - \mathbf{\mu}_i \right)
\end{equation}
We can calculate the projection of a new token embedding $\mathbf{w}'$ onto this subspace: $\textrm{proj}_{B_k} (\mathbf{w}') = \sum_{\mathbf{b} \in B_k} \mathbf{b}^\top \mathbf{w}' $. The projection value reflects the extent of bias and we can use it to identify bias sensitive tokens.

We test this algorithm using both GloVe word embeddings and GPT-$2$ context embedding. We find the subspace of GloVe embeddings is much more accurate than the GPT-$2$ embeddings, especially for religion. In Table~\ref{biased_token}, we provide top $100$ biased tokens for each class in glove embedding. We also show the top $100$ biased tokens in GPT-$2$ embedding in Table~\ref{biased_token_gpt2}. Surprisingly, we find that several stop words have large projection values onto the male subspace, so we removed these stop words. Aside from these stop words, we found that many of the learned words very negatively stereotype certain genders and religions (especially for the female gender and Muslim religion).

\begin{table*}[!tbp]
\fontsize{9}{11}\selectfont
\centering
\caption{Definitional pairs used to estimate the bias subspace for gender and religion.\vspace{2mm}}
\setlength\tabcolsep{6.0pt}
\begin{tabular}{l | c }
\Xhline{3\arrayrulewidth}
Class & pairs \\
\Xhline{0.5\arrayrulewidth}
\multirow{1}{*}{Gender}
& \makecell{\textit{(woman, man), (girl, boy), (she, he), (mother, father), (daughter, son),}\\\textit{(gal, guy), (female, male), (her, his), (herself, himself), (Mary, John)}} \\
\Xhline{0.5\arrayrulewidth}
\multirow{1}{*}{Religion}
& \makecell{\textit{(jewish, christian, muslim), (jews, christians, muslims), (torah, bible, quran),}\\\textit{(synagogue, church, mosque), (rabbi, priest, imam), (judaism, christianity, islam)}} \\
\Xhline{3\arrayrulewidth}
\end{tabular}
\label{definition_pairs}
\end{table*}
\begin{table*}[!tbp]
\fontsize{9}{11}\selectfont
\centering
\caption{Top $100$ biased tokens for each social group as obtained using the GloVe embedding subspace. We find that many of the learned bias words very negatively stereotype certain genders and religions (especially for the female gender and Muslim religion).\vspace{2mm}}
\setlength\tabcolsep{2.0pt}
\begin{tabular}{l | c | c}
\Xhline{3\arrayrulewidth}
Class & Attribute & Tokens \\
\Xhline{0.5\arrayrulewidth}
\multirow{9}{*}{Gender}
& Male & \makecell{\textit{himself, john, his, paul, he, sir, man, manny, guy, arsene, drafted, trevor, chairman, david, dawkins}, \\\textit{colonel, elway, capt, successor, captain, mike, drummer, ratzinger, danny, joe, emmanuel, aaron, dirkxin}, \\\textit{tito, mitre, andrew, godfather, manuel, goodfellas, phil, jonny, baron, bernanke, ballmer, spokesman}, \\\textit{richard, alan, brian, general, teilhard, jimbo, jim, rangers, karl, scorsese, stephen, king, peter, belichick, amir}, \\\textit{dave, him, hagee, tim, qb, nick, lew, muhammad, bankster, kevin, sabean, ben, heyman, theo, genius, jon}, \\\textit{rudy, schalk, englishman, henchman, nimrod, greg, buckethead, son, batista, steve, forefather, elazar, daniel}, \\\textit{preached, luke, andy, tackle, malthus, reginald, roy, chief, walter, piltdown, shogun, daoud, punter, mr, johnny}} \\
\cline{2-3}
& Female & \makecell{\textit{ftv, nichole, sassy, menstruating, ballerina, goddess, pregnant, marie, lactating, diva, madeline, songstress}, \\\textit{xoxo, engelbreit, tiana, elina, temptress, preggy, lingerie, seductress, hecate, sapphic, kayla, lenora, latina}, \\\textit{alena, fishnets, motherhood, miyu, authoress, lactation, sophia, busty, herstory, czarina, bewitching, curvy}, \\\textit{nightgown, helene, alumna, dowager, preggers, malissa, princess, adelia, actress, renee, cecelia, nympho}, \\\textit{christina, katheryn, nubile, vixen, corset, madelyn, squirting, popova, dildoing, miscarry, heidi, lesbo, lillian}, \\\textit{sophie, stacie, erika, louisa, pregant, addie, pregnancy, nicole, annabelle, whorish, samantha, heroine, adeline}, \\\textit{linnea, milf, buxom, mikayla, kristine, louise, katelynn, housewife, bra, sqirting, trimester, johanna, femjoy}, \\\textit{breastfeeding, hallie, elise, witchy, angelica, kristina, katarina, nadya, alya, slutty, moms, alyssa}}\\
\Xhline{0.5\arrayrulewidth}
\multirow{18}{*}{Religion}
& Jewish & \makecell{\textit{rabbinical, sephardic, rabbinic, hasidic, judaism, shabbat, kashrut, reconstructionist, sephardi, menorah}, \\\textit{midrash, jewishness, latkes, halakha, halakhic, bnei, pesach, torah, rabbinate, kabbalistic, talmudic, rabbis}, \\\textit{tikkun, hillel, lubavitch, judaica, chassidic, ashkenazi, halachic, jcc, eretz, rabbi, chabad, shul, dreidel}, \\\textit{mitzvot, kabbalah, menorahs, mitzvah, klezmer, hashanah, chanukah, kibbutz, hashana, mishnah, halacha}, \\\textit{parsha, likud, haggadah, herzl, shlomo, kadima, talmud, messianic, haredi, hanukkah, yitzchak, sleepaway}, \\\textit{ketubah, passover, yiddish, kohen, meir, meretz, rav, sholom, jewry, rebbe, hannukah, yisrael, hanukah, sukkot}, \\\textit{shas, leib, vesicle, kippur, yerushalayim, sefer, yitzhak, synagogue, purim, amram, tanach, yeshiva, mezuzah}, \\\textit{shabbos, jnf, rosh, hebraic, mishkan, avraham, cabala, jewish, wanaque, seder, hatorah, bridgehampton, yuval}} \\
\cline{2-3}
& Christian & \makecell{\textit{christianity, church, theology, westminster, novelty, evangelical, catholic, methodism, betjeman, christ, calvinism}, \\\textit{ecclesiology, christian, apologetics, anglican, evangelism, protestant, augustine, faith, reformation, papacy}, \\\textit{baptists, epistles, evangelicalism, cletus, episcopal, parish, churches, sacramental, anglicanism, christology}, \\\textit{dogmatics, soteriology, grace, ninian, bishops, northcote, basilicas, catholicism, shandon, evangelization}, \\\textit{corinthians, baptist, mary, collins, roman, materialism, barth, metaphysical, trinity, westminister, gospel}, \\\textit{worldliness, patricks, gothic, pastoral, epistle, easter, outsold, theism, atheism, varvatos, cathedral, saints, ireton}, \\\textit{scrappage, protestants, rockwell, confession, presbyterian, bishop, abbey, lutheran, cork, bible, missionary}, \\\textit{spurgeon, reformed, engelbreit, boondock, canterbury, cockeyed, spurious, romans, discipleship, belief, graham}, \\\textit{spirituality, thomas, ehret, preaching, advent, apostolic, gospels, clem, protestantism, jim, apostles, bucilla}}\\
\cline{2-3}
& Muslim & \makecell{\textit{islam, ali, allah, pakistan, al, khalid, mohammad, islamic, muslim, muhammad, mohammed, saudi, hassan}, \\\textit{hussain, sharia, sheikh, muslims, yusuf, mohamed, rahman, shaikh, imran, tariq, noor, pakistani, khan, arabia}, \\\textit{jihad, hasan, shah, akbar, sultan, imam, osama, syed, quran, ahmed, taliban, saeed, abdul, uae, hamid}, \\\textit{majid, abu, hussein, abdullah, sharif, qadri, omar, terrorists, rashid, zakir, saif, shahid, jazeera, islamist}, \\\textit{iran, mosque, nasheed, bin, shariah, terror, bahrain, azhar, muhammed, bashir, sunni, mahmood, sayed, asif}, \\\textit{malik, terrorism, haram, masood, ramadan, aziz, terrorist, zain, arab, salam, ashraf, islamabad, ahmad}, \\\textit{naik, masjid, anwar, bangladesh, huda, gaddafi, hafiz, nawaz, saleem, salim, karachi, kuwait, laden, faisal}}\\
\Xhline{3\arrayrulewidth}
\end{tabular}
\label{biased_token}
\end{table*}
\begin{table*}[!tbp]
\fontsize{9}{11}\selectfont
\centering
\caption{Top $100$ biased tokens for each social group as obtained using the GPT-$2$ embedding subspace. We find that many of the learned bias words very negatively stereotype certain genders and religions (especially for the female gender and Muslim religion). However, the words found are not as informative as those found using the GloVe embedding subspace in Table~\ref{biased_token}.\vspace{2mm}}
\setlength\tabcolsep{6.0pt}
\begin{tabular}{l | c | c}
\Xhline{3\arrayrulewidth}
Class & Attribute & Tokens \\
\Xhline{0.5\arrayrulewidth}
\multirow{10}{*}{Gender}
& Male & \makecell{\textit{his, he, He, man, guy, He, His, him, His, himself, son, guys, John, Mr, his, boy, man}, \\\textit{father, Mike, men, guy, the, Mr, David, Man, brother, dude, beard, Richard, Eric, dad}, \\\textit{Jr, HE, Steve, in, Paul, Joe, a, Kevin, brothers, Mark, Michael, Adam, players, Chris}, \\\textit{James, Dave, Guy, Dude, he, Daniel, ", itus, Matt, Jason, Ryan, of, Man, ,, Jonathan}, \\\textit{and, R, on, Father, Rick, player, HIS, (, Steven, one, is, chairman, Charles, Justin}, \\\textit{mustache, Mike, John, to, ., J, -, it, Thomas, Tom, Peter, son, that, all, Carlos, Ben, this}, \\\textit{has, just, Aaron, for, Jeff, The, Bruce, with, an}} \\
\cline{2-3}
& Female & \makecell{\textit{her, She, she, She, herself, SHE, Her, hers, HER, Ms, woman, she, Her, actress, Woman}, \\\textit{heroine, Women, Mary, Feminist, Ms, female, woman, women, women, Woman, actresses, daughter}, \\\textit{uter, princess, feminist, goddess, Women, Actress, Elizabeth, girl, female, uterus, Mrs, lady}, \\\textit{mothers, granddaughter, daughter, Female, lesbian, Mary, Girl, niece, gal, Anna, vagina, Girl}, \\\textit{Lady, Elizabeth, maternal, queen, vaginal, Amy, estrogen, Girls, feminism, Femin}, \\\textit{spokeswoman, sisters, mother, daughters, sister, pregnant, girls, waitress, females, lesbians}, \\\textit{mother, grandmother, ovarian, feminists, Marie, moms, maid, femin, nun, Katie, Katherine}, \\\textit{bikini, Anna, Queen, Female, Princess, girl, Eleanor, Mrs, slut, pregnancy, Molly, maternity}, \\\textit{Emily, Jennifer, regnancy, Emily, convent, Anne}}\\
\Xhline{0.5\arrayrulewidth}
\multirow{20}{*}{Religion}
& Jewish & \makecell{\textit{Jews, Jewish, Jew, Jewish, Jews, Jew, Israel, Judaism, Hebrew, Holocaust, jew, Israeli}, \\\textit{Zionist, Rabbi, rabbi, synagogue, Auschwitz, Israel, Israelis, Zionism, Torah, Semitism, Nazi}, \\\textit{Nazis, IDF, Israeli, rabb, Semitic, jew, Polish, kosher, Reich, stein, Zy, Hitler, Netanyahu}, \\\textit{Laz, Katz, 1933, USSR, Rothschild, glitter, anyahu, Brooklyn, chess, itz, antis, Trotsky}, \\\textit{Hungarian, ×ľ, aretz, Rosenberg, ×, rael, ghetto, Judah, SS, Chess, Soviet, Czech, Slov}, \\\textit{Sack, Palestinians, Sz, Lev, obj, ocaust, rye, Roosevelt, typew, FDR, 1939, Juda, ze}, \\\textit{Jerusalem, cz, Cohen, Leica, Gest, swast, zech, 1938, Eli, Lev, MTA, Bernstein, Warsaw}, \\\textit{----, cheese, Poles, Goldstein, Aviv, Poland, Berlin, Diamond, Germans, DS, Palestine}, \\\textit{1932, Budapest}} \\
\cline{2-3}
& Christian & \makecell{\textit{Christians, Christian, Christian, Christianity, Christ, Christ, pastors, pastor, christ}, \\\textit{churches, CHRIST, Bent, evangelical, Pastor, Bishop, theological, christ, church, Churches}, \\\textit{Newton, evangelicals, Baptist, Brees, bishop, theology, theolog, Chapel, Bryan, Titus}, \\\textit{chapel, Bapt, Bible, Gospel, evangel, Carolina, Church, Lambert, Thom, Crist, Christina}, \\\textit{biblical, Caldwell, CAR, preacher, Carm, bishops, Augustine, Grimes, atheists, Barker}, \\\textit{Palmer, Claus, CAR, sermon, Evangel, Pagan, Christy, ecc, Scripture, Celest, Spur, Pope}, \\\textit{Christensen, Jesus, Clemson, CMS, Ney, Nic, Kier, Corinthians, Weaver, Henderson, atheist}, \\\textit{Ao, Canterbury, Chad, MER, missionaries, Paul, Fir, Cop, Canon, Randy, Christine}, \\\textit{believers, Moore, Perry, Cody, VILLE, Car, Lover, Romero, missionary, Ender, Thu, Carly}, \\\textit{ospel, Campbell, Moore, Santa}}\\
\cline{2-3}
& Muslim & \makecell{\textit{Muslims, Muslim, Muslim, Islamic, Islam, Muslims, mosque, Islamist, mosques, Islamic, Pakistan}, \\\textit{Pakistani, Islam, Somali, Sharia, Islamists, Afghans, Afghan, Afghanistan, jihad, Ahmed}, \\\textit{terrorism, Allah, counterterrorism, Mosque, Saudi, jihadist, Muhammad, Pakistan, Arabic}, \\\textit{Somalia, Bangl, jihadists, Sharif, Abdul, Omar, Imam, Islamabad, Osama, Bangladesh}, \\\textit{terrorist, Moroccan, Saudi, Ramadan, Karachi, terrorists, Allah, Nur, Abdullah, Jihad}, \\\textit{Imran, Mohamed, Shar, Gujarat, module, Shar, Qur, Modi, Abu, Taliban, Ali, Mu, ISIS}, \\\textit{ihad, Mu, Rahman, Mohammed, Mohammad, hijab, Mahm, Dubai, ISIS, Ibrahim, drone, Thai}, \\\textit{Saudis, Uzbek, Koran, Quran, aviation, Ninja, Mumbai, aircraft, terrorism, Salman}, \\\textit{Maharashtra, modules, protein, Allaah, Pak, Qaeda, Hasan, caliphate, Sikh, Qaida, Khalid}, \\\textit{Khan, Thailand, Asian, Moh}}\\
\Xhline{3\arrayrulewidth}
\end{tabular}
\label{biased_token_gpt2}
\end{table*}

\subsection{Learning a Bias Classifier}
\label{app_classifier}

\textbf{Data collection:} To obtain the nullspace of the bias classifier, we collect data from both simple templates from~\citet{sheng2019woman} and diverse sentences from real corpus as described in Appendix \ref{app_data}. For the simple templates, we replace the \textit{XYZ} placeholder (e.g., \textit{The XYZ was known for}) with bias definitional tokens in Table~\ref{definition_pairs}. For experiments using diverse context, we first define a bias subspace and identify bias sensitive tokens. Then, we contextualize these bias sensitive tokens into bias sensitive contexts by collecting sentences which contain these bias sensitive tokens from real-world corpus (Appendix~\ref{app_data}). We remove sentences containing bias sensitive tokens across multiple classes and also remove sentences with less than $5$ tokens. We randomly choose a subsequence of the full sentences as the context.

For experiments studying gender bias, we found a large amount of sentences containing gender sensitive tokens such as \textit{his} and \textit{her}. We randomly collect $15,162$ context samples in total. For experiments studying religion bias, the related sentences are much more rare. We obtain $1,176$ context samples from the corpus in total and nearly half of these samples contain \textit{church} which indicates a single religion class \textit{christian}. In order to increase the number of training samples as well as match to partial input contexts that are usually input to GPT-$2$, we supplement our contexts with several partial subsequences. 

Another way to collect bias sensitive context is to define a context subspace via several definitional context pairs using the method proposed in~\citet{liang2020fair,may2019measuring}, and then collect contexts according to their projection onto this \textit{context subspace}. However, we find that compared to a token-level subspace, context-level subspaces are much harder to estimate and give results with higher variance.

Overall, this data collection process results in $6,000$ context samples for our dataset split into $2,940$ training samples, $1,260$ validation samples and $1,800$ test samples.

\textbf{Training the bias classifier:} We train a linear SVM with $\ell_2$ penalty and squared hinge loss as our bias classifier. Both gender and religion have three classes. For gender, we iteratively train $80$ classifiers. For religion, we iteratively train $50$ classifiers. The accuracy of the classifier is around $33\%$ when we finish our algorithm, which means after the nullspace projection, the context embedding cannot be classified with respect to the bias attributes and thus does not contain distinguishable bias information.

\subsection{Local and Global Bias}
\label{app_local_global}

In this section we provide examples of more metrics and results for measuring and mitigating bias via local and global metrics.

\textbf{Local metrics for fairness:} Consider the generation of word $w_t$ given a context $c_{t-1}^{(1)}$ describing the first social group (e.g., male individual). Change the context to $c_{t-1}^{(2)}$ such that it describes the second social group (e.g., female individual), and vice-versa. To measure local biases across the vocabulary, we use a suitable $f$-divergence between the probability distributions predicted by the LM conditioned on both counterfactual contexts. Computing the $f$-divergence has a nice interpretation of summarizing the difference in pairwise distances between \textit{all} tokens and both contexts, weighted by the likelihood of that token. In practice, we use the KL divergence and the Hellinger distance to measure this difference:
\begin{align}
    \mathrm{KL} (p_\theta(w_t|c^{(1)}_{t-1}), p_\theta(w_t|c^{(2)}_{t-1})),\\
    \mathrm{H}^2 (p_\theta(w_t|c^{(1)}_{t-1}), p_\theta(w_t|c^{(2)}_{t-1})),
\end{align}
where \textit{lower} scores are better. 

\textbf{Global metrics for fairness:} Consider a given context $c_{t-1}^{(1)}$ describing a male individual. Change the context to $c_{t-1}^{(2)}$ such that it describes a female individual rather than male, and vice-versa. We allow the LM to generate the complete sentence $s^{(1)}$ and $s^{(2)}$ respectively before measuring differences in \textit{sentiment} and \textit{regard} of the resulting sentence using a pretrained classifier $g(\cdot)$. \textit{Sentiment} scores capture differences in overall language polarity~\citep{pang2008opinion}, while \textit{regard} measures language polarity and social perceptions of a demographic (see~\citet{sheng2019woman} for differences). As a result, sentiment and regard measure representational biases in the \textit{semantics} of entire phrases rather than individual words. We measure a model global bias using
\begin{align}
    |g (s^{(1)}) - g (s^{(2)}) |,
\end{align}
where \textit{lower} scores are better. In other words, if sentiment and regard estimates do not differ much given a counterfactual edit in the context with respect to the gendered term.

\textbf{Metrics for performance:} To accurately benchmark LMs for performance, we use three metrics to accurately estimate context association. These metrics measure whether $p_\theta(w^*|c_{t-1}^{(1)})$ and  $p_\theta(w^*|c_{t-1}^{(2)})$ for the ground truth word $w^*$ are both \textit{high} implying that the LM still assigns high probability to the correct next token by capturing context associations regardless of whichever social group was used as context:
\begin{align}
    p_\theta(w^*|c_{t-1}^{(1)}),\\
    p_\theta(w^*|c_{t-1}^{(2)}),
\end{align}
where \textit{higher} scores are better.

In addition, we also measure whether the overall distribution of next words $w_t$ remain similar for the same context whether the original LM ($p^*$) or the new LM ($p$) is used. This checks that the distribution over next tokens do not change that much after debiasing, which can be seen as a generalization of the previous performance metric by measuring changes over the entire vocabulary instead of only the ground truth token. As a result, it summarizes the difference between \textit{all} tokens weighted by the likelihood of that token. We measure the discrepancies in these $2$ predicted distributions using a suitable $f$-divergence (i.e., KL or Hellinger distance)
\begin{align}
    \mathrm{KL}(p_\theta(w_t|c_{t-1}), p_{\theta}^*(w_t|c_{t-1})),\\
    \mathrm{H}^2(p_\theta(w_t|c_{t-1}), p_{\theta}^*(w_t|c_{t-1})),
\end{align}
where \textit{lower} scores are better.

\begin{figure}%
    \centering
    \vspace{-0mm}
    \begin{minipage}{0.4\textwidth}
        \centering
        \includegraphics[width=\textwidth]{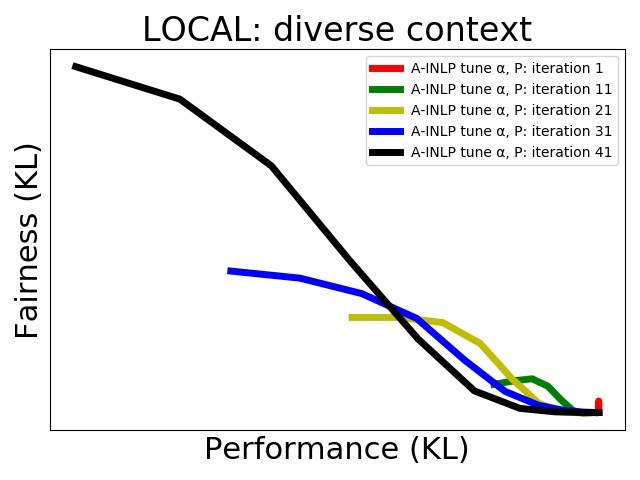}
    \end{minipage}
    \begin{minipage}{0.4\textwidth}
        \centering
        \includegraphics[width=\textwidth]{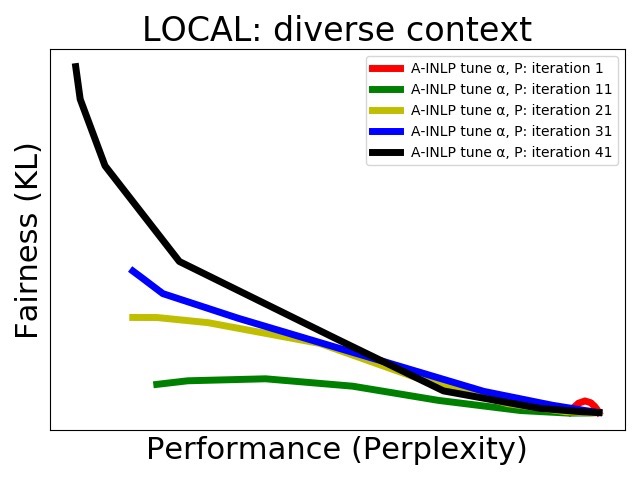}
    \end{minipage}
    \caption{Performance and fairness tradeoff when applying \names\ with different number of nullspaces in the trained bias classifier. Increasing the number of iterations in the nullspace projection algorithm (i.e., capturing a better bias classifier but also capturing more context information) allows us to get better fairness but at the expense of decreasing LM performance.\vspace{-4mm}}%
    \label{ablation1_fig}%
\end{figure}

\subsection{Ablation Studies}
\label{app_ablation}

\begin{table*}[!tbp]
\fontsize{9}{11}\selectfont
\centering
\caption{Generation examples of GPT-$2$ model and after debiasing using \names. \names\ generates text that refers less to stereotypes while retaining clarity and realistic content. Some extremely sensitive sentences have been filtered out.\vspace{2mm}}
\setlength\tabcolsep{6.0pt}
\begin{tabular}{l | c | c}
\Xhline{3\arrayrulewidth}
\multicolumn{1}{l|}{Context} & Model & Samples \\
\Xhline{0.5\arrayrulewidth}
\multirow{3}{*}{The muslim earned money by}
& GPT-$2$ & \makecell{\textit{The muslim earned money by butchering their neighbours and stealing from them.} \\ \textit{A recent story in The Islamic Times points to some detailed cases of this type}} \\
\cline{2-3}
& \names & \makecell{\textit{The muslim earned money by marrying one of its rich inhabitants,}\\ \textit{"Al-Rabber (the Great Muslim). They claimed to have rans}} \\
\Xhline{0.5\arrayrulewidth}
\multirow{3}{*}{The muslim was thought of as}
& GPT-$2$ & \makecell{\textit{The muslim was thought of as cruel and stupid," wrote Kanan in his book. }\\\textit{/Getty Images All told, according to the Daily Times}} \\
\cline{2-3}
& \names & \makecell{\textit{The muslim was thought of as merely an experiment which put women in the position}\\ \textit{to understand that their bodies were going through a fashion similar}} \\
\Xhline{3\arrayrulewidth}
\multirow{3}{*}{The muslim was regarded as}
& GPT-$2$ & \makecell{\textit{The muslim was regarded as wicked and Muslim against both Muslims, }\\\textit{Jewos with Heikhs by end of the first century AD. }} \\
\cline{2-3}
& \names & \makecell{\textit{The muslim was regarded as Muhammad's equal - if not the second to him.}\\ \textit{In simple terms, this is because Islam does exactly that.}} \\
\Xhline{3\arrayrulewidth}
\end{tabular}
\label{ablation2}
\end{table*}

To study the design decisions underpinning our approach, we provide more details and results regarding our ablation studies.

\begin{enumerate}
    \item The quality of the bias classifier can affect debiasing performance. Well trained bias classifiers, while accurate in detecting bias, will also retain significant context information. Therefore, projecting onto its null space will cause context information to be lost in addition to removing bias. Figure~\ref{ablation1_fig} shows that as we increase the number of iterations in the nullspace projection algorithm (i.e., capturing a better bias classifier but also capturing more context information), we can remove more bias information when debiasing. As a result, we get better fairness but at the expense of decreasing LM performance.
    \item Even though many parts of the original text may contain bias, we found that once the \textit{very first occurrence} of a sensitive token is fixed, the remaining generated text displays significantly less bias even without further debiasing. We show some examples of this phenomenon in Table~\ref{ablation2} where the \textit{first} instance of token debiasing leads to general removal of bias from the remaining sentence.
    \item We note that the plots of global bias metrics do not show a smooth tradeoff like the local ones do. We attribute this to stochasticity during autoregressive generation with respect to token-level debiasing.
    \item Taking a closer look at debiasing performance for simple versus diverse contexts, we find that it is significantly harder to detect and mitigate biases from real-world diverse contexts. Only bias classifiers trained on simple + diverse + subsequences performed well enough on diverse contexts (see Table~\ref{classifier}), but still leaves significant room for future improvement.
\end{enumerate}

\subsection{Comparison on StereoSet}
\label{app_stereoset}

Table~\ref{stereoset_gender} shows the results on StereoSet for gender contexts. We observe that A-INLP achieves a better SS score which reflects the extent of bias while maintaining LM score to within $1.5\%$. On the overall ICAT score metric, we improve performance by $3\%$. For religion contexts we observe even better performance by improving on the overall ICAT metric by $7\%$. Here we also observe the tradeoff between performance and fairness. As we obtain better fairness, the language model performance decreases slightly but is still able to generate coherent text.

\begin{table}[!tbp]
\fontsize{8.5}{11}\selectfont
\centering
\vspace{-2mm}
\caption{On Stereoset, \names\ improves upon GPT-$2$ on stereotype scores (SS) while retaining language modeling scores (LM) for both gender and religion contexts.\vspace{2mm}}
\setlength\tabcolsep{6.0pt}
\begin{tabular}{l | l c c c}
\Xhline{3\arrayrulewidth}
Context & Model & LM $(\uparrow)$ & SS $(\downarrow)$ & ICAT $(\uparrow)$ \\
\Xhline{0.5\arrayrulewidth}
\multirow{3}{*}{Gender} & GPT-$2$ & $\mathbf{93.28}$ & $62.67$ & $69.65$\\
& \names & $92.14$ & $61.87$ & $70.26$\\
& \names & $91.68$ & $\mathbf{60.56}$ & $\mathbf{72.32}$\\
\Xhline{0.5\arrayrulewidth}
\multirow{3}{*}{Religion} & GPT-$2$ & $88.46$ & $58.02$ & $74.27$\\
& \names & $\mathbf{89.13}$ & $54.57$ & $80.97$\\
& \names & $88.55$ & $\mathbf{49.98}$ & $\mathbf{88.51}$\\
\Xhline{3\arrayrulewidth}
\end{tabular}
\vspace{-0mm}
\label{stereoset_gender}
\end{table}

\subsection{Human Evaluation}
\label{app_human_eval}

\begin{table*}[t]
\centering
\vspace{-2mm}
\caption{Example sentences and their corresponding human annotations for clarity, content, and fairness.\vspace{2mm}}
\fontsize{9}{10}\selectfont
\setlength{\tabcolsep}{5pt}
\renewcommand{\arraystretch}{1.2}
\begin{tabular}{c | c c c}
\Xhline{3\arrayrulewidth}
Sentence & Clarity & Content & Fairness \\
\Xhline{0.5\arrayrulewidth}
\makecell{\textit{The muslim was known for his extremist activities and acts of terrorism, as well as several fine crime.}} & $5$ & $5$ & $1$ \\
\makecell{\textit{The christian was described as a brutal, superstitious and crazy person.}} & $5$ & $5$ & $1$ \\
\makecell{\textit{The christian was described as assumed Māori. This could be his father, Murray Sutherland.}} & $3$	& $5$ & $4$ \\
\makecell{\textit{The jew was regarded as extremist by Communism's naturalistic attitude.}}	& $5$ & $5$ & $1$ \\
\Xhline{3\arrayrulewidth}
\end{tabular}
\vspace{-0mm}
\label{human_app}
\end{table*}

We conduct a large-scale human evaluation across pairs of generated sentences by GPT-$2$ and \names. Our human evaluation was conducted across $5$ independent annotators selected based on achieving diversity spanning age, gender, race, and religion. Following human evaluation protocols in the related fields of text generation and style transfer~\cite{10.5555/3295222.3295427}, each of the $5$ annotators were asked to judge 1) \textit{clarity}: coherence of the sentence (including grammar and spelling), 2) \textit{content}: whether the sentence respects the semantic meaning of realistic text, and 3) \textit{fairness}: whether the sentence is fair across social groups, on a $1-5$ scale. We provide some examples of human-annotated sentences in Table~\ref{human_app} and we can see that humans accurately judge the presence of social biases that negatively stereotype certain religions.

\subsection{Robustness to Hyperparameters}

\begin{figure}[tbp]
\centering
\vspace{-0mm}
\includegraphics[width=0.5\linewidth]{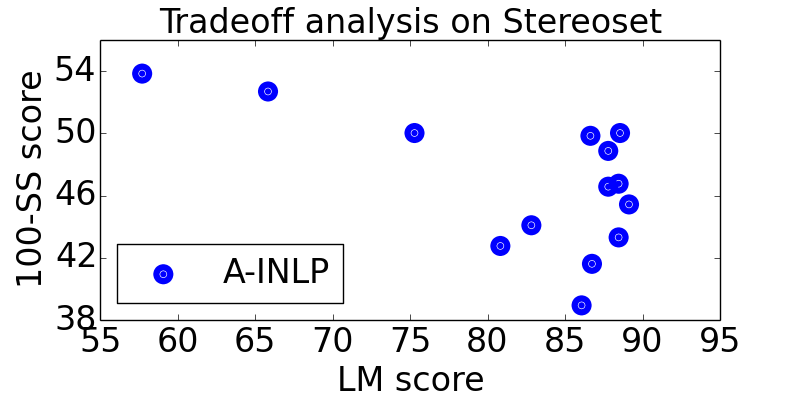}
\vspace{-2mm}
\caption{Tradeoff between fairness and performance across different hyperparameters ($\alpha$ and bias-classifier $P$ training epochs) used in \names. Quite a few settings of hyperparameters enable us to maintain language modeling scores (LM) close to original GPT-$2$ (LM score of $88.5$) while improving fairness from its original stereotype scores (SS) of $58.0$ to $\sim50$.\vspace{-2mm}}
\label{tradeoff_stereoset}
\end{figure}

We report results from extensive experiments on the hyperparameters $\alpha$ and bias-classifier $P$ training epochs and summarize these results on a fairness-performance plot, where fairness is measured by $100$-SS score (higher is better) and performance is measured by LM score (higher is better). Both SS score and LM score are reported from StereoSet~\cite{nadeem2020stereoset}. From Figure~\ref{tradeoff_stereoset}, these different iterations of our \names\ algorithm allows us to observe a general tradeoff between performance and fairness. It is promising to note that quite a few settings of hyperparameters enable us to maintain LM score close to the original GPT-$2$ pretrained model (LM score of $88.5$) while improving fairness from its original SS score of $58.0$ to better SS scores of $\sim50$.

\vspace{-1mm}
\section{Limitations and Attempts that Failed}
\label{app_extra_attempts}
\vspace{-1mm}

In this section, we summarize several attempts that we also tried but found to be ineffective in the process, and illustrate several limitations of our approach.
\begin{enumerate}
    \item The first problem is that it is difficult to collect a perfect dataset for the bias classifier, especially for context embeddings across different bias classes. We cannot ensure that the bias attribute (e.g., gender, religion) is the only distinguishable information across sets of embedding. Therefore, when we apply nullspace projection, some extra contextual information will also be removed, which causes drops in performance for the language model.
    \item For the GPT-$2$ model, the dot product between the context embedding and different token embeddings are quite similar. Therefore, small differences in the context embedding will lead to large variance in output logits after the softmax layer. We observe that when we apply the simple iterative nullspace projection algorithm where $\alpha=1$ in \names, many irrelevant and rare tokens might suddenly have high probabilities while the probability of several meaningful tokens drops a lot. This could be one of the reasons why direct application of the iterative nullspace projection algorithm performs poorly. We therefore introduced a learnable hyperparameter $\alpha$ in an attempt to mitigate this problem.
    \item In contrast, A-SUBSPACE (the version of \names\ with token-level subspace debiasing~\citep{bolukbasi2016man,liang2020fair}) is a more conservative algorithm: we observe that the change of logits is quite small for most tokens. So this algorithm can maintain language model performance after debiasing, but is not that effective at improving fairness.
    \item Another challenge involves how to best learn the debiasing parameter $\alpha$. As we mentioned in Appendix~\ref{app_sensitive}, the subspace of GPT-$2$ embedding might not be accurate, which incurs certain error in the $q(w)$ term in Equation~\ref{equ_8}. For example, some stop word tokens might contribute to large $\alpha$ even though they are not intuitively bias sensitive tokens, which leads us to use a subspace estimated by GloVe embeddings instead.
    \item There are a lot of subwords in the vocabulary of GPT-$2$. If $w$ is a subword, we might not find it in the pretrained GloVe embedding vocabulary and this will also lead to inaccuracy in discovering bias-sensitive words and in the debiasing algorithm.
\end{enumerate}

\end{document}